\pdfoutput=1

\documentclass[11pt]{article}

\usepackage[]{acl}

\usepackage{times}
\usepackage{latexsym}

\usepackage[T1]{fontenc}

\usepackage[utf8]{inputenc}

\usepackage{microtype}
\usepackage{amsmath}
\usepackage{url}
\usepackage{graphicx}
\usepackage{graphics}
\usepackage{enumitem}
\usepackage{subcaption}
\usepackage{caption}
\usepackage{amssymb}
\usepackage{verbatim}
\usepackage[normalem]{ulem} 
\usepackage{makecell}
\usepackage{array}
\usepackage{url}
\usepackage{comment}
\usepackage{multirow}
\usepackage{gensymb}
\usepackage{float}
\usepackage{bbm}
\usepackage{dsfont}
\usepackage{pgfplots}
\usepackage{layouts}

\newlength\figureheight
\newlength\figurewidth
\newcommand\inputpgf[2]{{
\let\pgfimageWithoutPath\pgfimage
\renewcommand{\pgfimage}[2][]{\pgfimageWithoutPath[##1]{#1/##2}}
\input{#1/#2}
}}

\title{Active Evaluation: Efficient NLG Evaluation with Few Pairwise Comparisons}

\author{Akash Kumar Mohankumar\thanks{* Work done at Indian Institute of Technology Madras}\\
  Microsoft \\
  Bangalore, India \\
  \texttt{makashkumar@microsoft.com} \\\And
  Mitesh M. Khapra \\
  Indian Institute of Technology Madras \\
  RBCDSAI, IIT Madras \\
  \texttt{miteshk@cse.iitm.ac.in} \\}

\begin{document}
\maketitle
\begin{abstract}
Recent studies have shown the advantages of evaluating NLG systems using pairwise comparisons as opposed to direct assessment. Given $k$ systems, a naive approach for identifying the top-ranked system would be to uniformly obtain pairwise comparisons from all ${k \choose 2}$ pairs of systems. However, this can be very expensive as the number of human annotations required would grow quadratically with $k$. In this work, we introduce \textit{Active Evaluation}, a framework to efficiently identify the top-ranked system by actively choosing system pairs for comparison using dueling bandit algorithms. We perform extensive experiments with 13 dueling bandits algorithms on 13 NLG evaluation datasets spanning 5 tasks and show that the number of human annotations can be reduced by 80\%. To further reduce the number of human annotations, we propose model-based dueling bandit algorithms which combine automatic evaluation metrics with human evaluations. Specifically, we eliminate sub-optimal systems even before the human annotation process and perform human evaluations only on test examples where the automatic metric is highly uncertain. This reduces the number of human annotations required \textit{further} by 89\%. In effect, we show that identifying the top-ranked system requires only a few hundred human annotations, which grow linearly with $k$. Lastly, we provide practical recommendations and best practices to identify the top-ranked system efficiently. Our code has been made publicly available at \url{https://github.com/akashkm99/duelnlg}
\end{abstract}

\section{Introduction}
In the last few years, the field of NLG has made rapid progress with the advent of large-scale models trained on massive amounts of data \cite{transformer, mT5, MBART, GPT3}. However, evaluation of NLG systems continues to be a challenge. On the one hand, we have automatic evaluation metrics which are easy to compute but unreliable. In particular, many studies have shown that they do not correlate well with human judgments \cite{NovikovaDCR17, ElliottK14, Ananya-ADEM, DEB, nlg_survey}. On the other hand, we have human evaluations, which are relatively more reliable but tedious, expensive, and time-consuming. Further, recent studies have highlighted some limitations of human evaluations that involve direct assessment on an absolute scale, \textit{e.g.}, Likert scale. Specifically, human evaluations using direct assessment have been shown to suffer from \textit{annotator bias}, \textit{high variance} and \textit{sequence effects} where the annotation of one item is influenced by preceding items \cite{Kulikov2019ImportanceOS, Sudoh2021IsTT, Liang2020BeyondUS, See2019WhatMA,Mathur2017SequenceEI}. 

In this work, we focus on reducing the cost and time required for human evaluations while not compromising on reliability. We take motivation from studies which show that selecting the better of two options is much easier for human annotators than providing an absolute score, which requires annotators to maintain a consistent standard across samples \cite{Kendall1948RankCM, SimpsonG18}. In particular, recent works show that ranking NLG systems using pairwise comparisons is a more reliable alternative than using direct assessment \cite{See2019WhatMA,Li2019ACUTEEVALID, Sedoc2019ChatEvalAT, Dhingra2019HandlingDR}. While this is promising, a naive approach for identifying the top-ranked system from a set of $k$ systems using uniform exploration is prohibitively expensive. Specifically, uniform exploration obtains an equal number of annotations for all the $k \choose 2$ system pairs; as a result, the required human annotations grows as $O(k^2)$. 

To reduce the number of pairwise annotations, we introduce Active Evaluation, a framework to efficiently identify the top-ranked NLG system. Our Active Evaluation framework consists of a learner that selects a pair of systems to compare at each time step. The learner, then, receives a feedback signal indicating the (human) preference between the selected systems on one input context, randomly sampled from the test dataset. The learner's objective is to reliably compute the top-ranked system with as few human annotations as possible. We adopt algorithms from the stochastic dueling bandits literature \cite{dueling_bandits_survey} to decide which pair of NLG systems to compare at each time step. To check if existing dueling bandits algorithms can indeed provide reliable top-rank estimates with minimal annotations, we evaluate 13 such algorithms on 13 NLG evaluation datasets spanning five tasks \textit{viz.}, machine translation, summarization, data-to-text generation, paraphrase generation, and grammatical error correction. We show that the best performing dueling bandit algorithm can reduce the number of human annotations by 80\% when compared to uniform exploration. 

To further reduce human annotations, we leverage automatic evaluation metrics in our Active Evaluation framework. We utilize existing automatic metrics such as BLEU \cite{bleu}, BertScore \cite{bertscore}, \textit{etc} for pairwise evaluations by converting the direct evaluation scores into preference probabilities using pairwise probability models. We also develop trained pairwise metrics that directly predict the comparison outcome given pairs of generated texts and context or reference as input. To incorporate such evaluation metrics in our Active Evaluation framework, we propose three model-based dueling bandits algorithms, \textit{viz.}, (i) Random Mixing: human annotations and evaluation metric predictions are randomly mixed, (ii) Uncertainty-aware selection: human annotations are obtained only when the predictions from the evaluation metric is highly uncertain, (iii) UCB Elimination: poorly performing NLG systems are eliminated using an Upper Confidence Bound (UCB) on the evaluation metric scores. Through our experiments, we show that the number of human annotations can be further reduced by 89\% on average (this reduction is over and above the 80\% reduction that we got earlier). In effect, we show that given $k$ systems, we can find the top-ranked NLG system efficiently with just a few hundred comparisons that vary as $O(k)$. Lastly, we provide practical recommendations to efficiently identify the top-ranked NLG system based on our empirical study on various design choices and hyperparameters.

\section{Active Evaluation Framework}
\label{sec:online_learning_framework}
We introduce the problem and our Active Evaluation setup in section \ref{subsec:problem_formulation}. Later in section \ref{subsec:choosing_system_pairs}, we describe the different approaches to decide which pairs of NLG systems to compare at each time step. Finally, in section \ref{subsec:top_rank_system}, we formalize the notion of top-ranked system. 
\subsection{Problem Formulation and Setup}
\label{subsec:problem_formulation}
\begin{figure}
    \centering
    \includegraphics[width=0.7\linewidth]{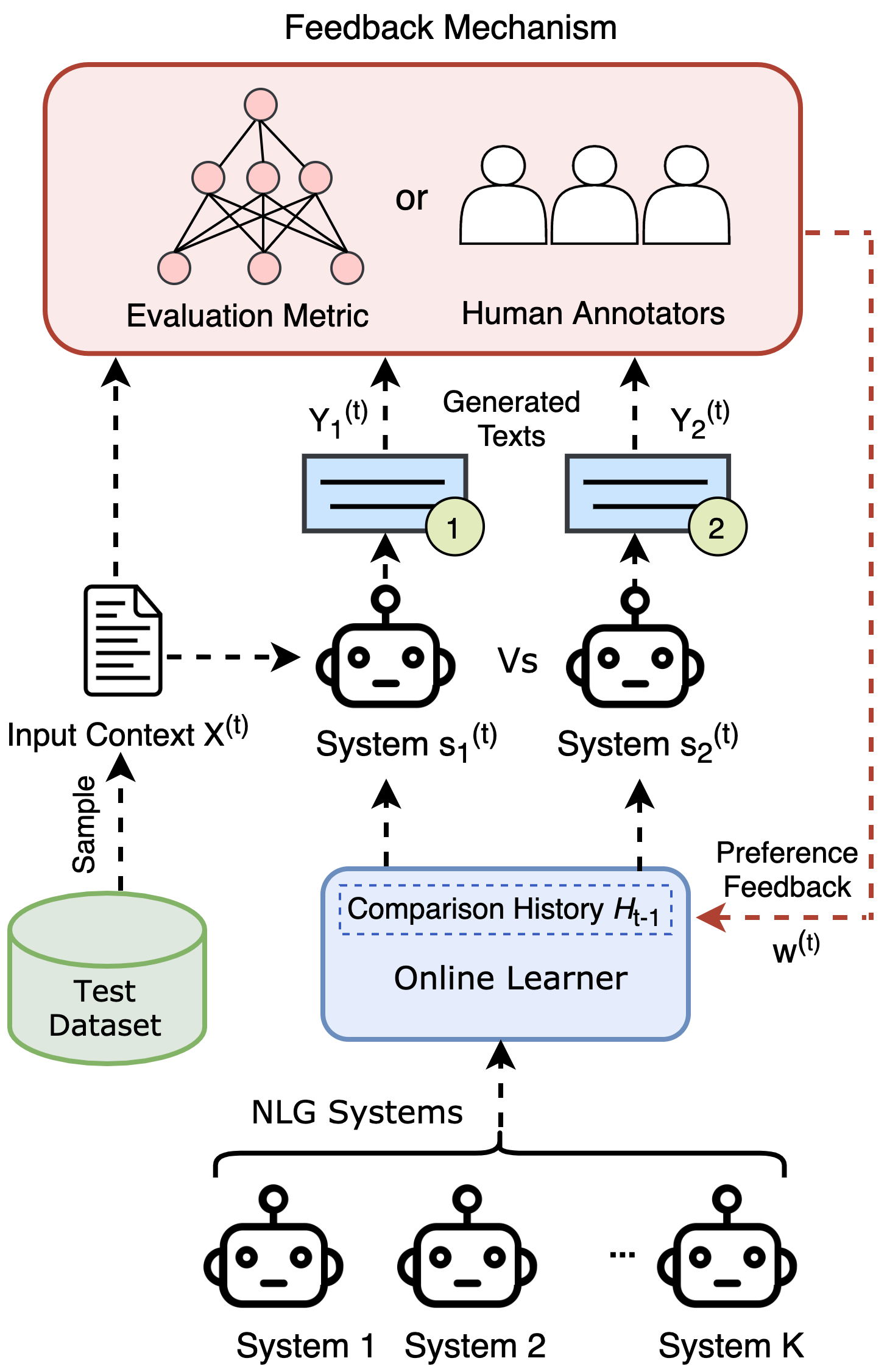}
    \caption{Our Active Evaluation framework consisting of a learner that chooses a pair of systems to compare at each time step. The learner receives feedback from either human annotators or the automatic metric.}
    \label{fig:duelling_nlg}
\end{figure}

We consider the problem of finding the top-ranked NLG system from a given set of $k$ systems, denoted by $\mathcal{S} = \{1, 2, \dots, k\}$. Our Active Evaluation framework consist of a \textit{learner} which at each time step $t$, chooses a pair of systems $s^{(t)}_1, s^{(t)}_2 \in \mathcal{S}$ for comparison. Then, we ask human annotators to compare the outputs of the chosen systems on a randomly sampled input context and provide the comparison outcome as feedback to the learner. Specifically, we first sample an input context $X^{(t)}$ from the test dataset and obtain the generated texts ${Y}^{(t)}_{1}, {Y}^{(t)}_{2}$ from the chosen systems $s^{(t)}_1, s^{(t)}_2$. We then display the generated texts ${Y}^{(t)}_{1}, {Y}^{(t)}_{2}$ along with the context $X^{(t)}$ to human annotators and obtain a comparison outcome $w^{(t)} = 1, 0$, or $0.5$ denoting whether ${Y}^{(t)}_{1}$ is of better, worse, or equal (tie) quality as  ${Y}^{(t)}_{2}$. Note that the feedback $w^{(t)}$ indicates the preference on only one input sample and not the entire test dataset. The overall framework is depicted in figure \ref{fig:duelling_nlg}. The learner's objective is to find the top-ranked system with as few pairwise comparisons as possible.
\subsection{Choosing System Pairs for Comparison}
\label{subsec:choosing_system_pairs}
The learner should decide the pair of systems $(s^{(t)}_1, s^{(t)}_2)$ to compare at each time step $t$. The naive approach is to uniformly explore all the ${k \choose 2}$ system pairs. Specifically, the probability of selecting a pair $(i,j), i\ne j$ at time $t$ is given by 
\begin{align*}
    P_{uniform}((s^{(t)}_1, s^{(t)}_2) = (i,j)) = \frac{1}{{k \choose 2}}
\end{align*}
However, as we show in our experiments, the number of human annotations required to find the top-ranked system by this approach is very expensive and grows quadratically with the number of systems since we equally explore all ${k \choose 2}$ pairs. To reduce the number of annotations, we use dueling bandit algorithms to actively choose pairs of systems to compare based on the history of previous observations. We provide an overview of 13 dueling bandits algorithms proposed in the literature in appendix \ref{appendix:summary_dueling_bandits}. We refer the readers to \cite{dueling_bandits_survey} for a complete survey. 
\subsection{Identifying the top-ranked system}
\label{subsec:top_rank_system}
We now formalize the notion of the top-ranked system. Let $p_{ij}$ denote the preference probability of system $i$ over system $j$ \textit{i.e.} the probability that a generated text from system $i$ is preferred over system $j$ in the test dataset. We say that a system $i$ "beats" system $j$ if $p_{ij} > \frac{1}{2}$. In other words, system $i$ beats system $j$ if the probability of winning in a pairwise comparison is larger for $i$ than it is for $j$. We define the top-ranked system $i^*$ as the one that beats all other systems, \textit{i.e.} $p_{i^*j} > \frac{1}{2}, \forall j \in \mathcal{S} - i^*$. 
\section{Pairwise Probability Models}
\label{sec:pairwise_probability_model}
Our Active Evaluation framework, which we described in the previous section, completely relied on human annotators to compare pairs of generated texts $({Y}_{1}, {Y}_{2})$ to provide the preference feedback $w$. We can further reduce the number of required human annotations by estimating the human preference feedback using automatic evaluation metrics. However, most existing evaluation metrics are designed for direct assessment and not directly suitable for pairwise evaluations. In this section, we describe three pairwise probability models to convert direct evaluation scores into pairwise preference probabilities. Let $f({Y})$ denote the score provided by a direct assessment metric $f$ to a generated text ${Y}$ (The dependence of $f$ on the reference/context is omitted for brevity). The pairwise preference probability $\hat{p}({Y}_{1} \succ {Y}_{2})$ between any two hypotheses ${Y}_{1}$ and ${Y}_{2}$ can be modeled in 3 different ways:
\begin{itemize}
    \item \textbf{Linear:}
    $$\hat{p}({Y}_{1} \succ {Y}_{2}) = \frac{1}{2} + (f({Y}_{1})- f({Y}_{2}))$$ 
    \item \textbf{Bradley-Terry-Luce (BTL)} \cite{Bradley1952RankAO, Luce1979IndividualCB}:
    $$\hat{p}({Y}_{1} \succ {Y}_{2}) = \frac{f({Y}_{1})}{f({Y}_{1}) + f({Y}_{2})}$$ 
    \item \textbf{BTL-logistic:} $$\hat{p}({Y}_{1} \succ {Y}_{2}) =\frac{1}{1 + e^{(f(Y_1) - f(Y_2))}}$$
\end{itemize}

As detailed in appendix \ref{appendix:pairwise_probability_models}, we appropriately preprocess the scores $f(Y)$ to ensure that preference probability lies between 0 and 1. We can now predict the comparison outcome $w$ by thresholding the preference probability at two thresholds $\tau_1$ and $\tau_2 (\geq \tau_1)$ to incorporate ties \textit{i.e.}:
\[
    \hat{w}= 
\begin{cases}
    1, & \text{if  } \hat{p}(Y_1 \succ Y_2) > \tau_2\\
    0, & \text{if  } \hat{p}(Y_1 \succ Y_2) < \tau_1\\
    0.5,              & \text{Otherwise}  
\end{cases}
\]
We choose $\tau_1$ and $\tau_2$ using grid search on the validation set. Refer appendix   \ref{appendix:pairwise_probability_models} for more details.

\section{Model-based Dueling Bandits}
In the previous section, we discussed pairwise probability models to obtain the estimated preference probability $\hat{p}(Y_1 \succ Y_2)$ and the comparison outcome $\hat{w}$ using scores assigned by direct assessment metrics. We now propose three model-based dueling bandit algorithms wherein we combine such predictions from evaluation metrics with human annotations in the Active Evaluation framework.
\label{sec:model_based}
\subsection{Random Mixing}
Here, we randomly provide either the real (human) or the evaluation metric predicted feedback to the learner. Specifically, at any time $t$, we use the predicted comparison outcome $\hat{w}^{(t)}$ as the feedback with probability $p_{m}$ and use human annotations ${w}^{(t)}$ as feedback with probability $1-p_{m}$. The hyperparameter $p_{m}$ controls the ratio of estimated and real feedback given to the learner. As with other hyperparameters, we tune $p_{m}$ on the validation set. 
\subsection{Uncertainty-aware Selection}
\label{subsec:uncertainty_aware}
In this algorithm, we estimate uncertainty in the evaluation metric predictions and decide to ask for human annotations only when the evaluation metric is highly uncertain. We specifically focus on trainable neural evaluation metrics such as Bleurt \cite{bleurt} where we estimate the prediction uncertainty using recent advances in Bayesian deep learning. Let $\hat{p}(Y_1 \succ Y_2 | \theta)$ denote the preference probability modelled by a neural evaluation metric with parameters $\theta$. Given a training dataset $\mathcal{D}^{tr}$, Bayesian inference involves computing the posterior distribution $p(\theta | \mathcal{D}^{tr})$ and marginalization over the parameters $\theta$: 
\begin{align*}
    \hat{p}(Y_1 \succ Y_2 |  \mathcal{D}^{tr}) = \int_{\theta} &\hat{p}(Y_1 \succ Y_2 | \theta) \hat{p}(\theta | \mathcal{D}^{tr})d\theta
\end{align*}
However, computing the true posterior and averaging over all possible parameters is intractable in practice. Hence, several approximations have been proposed in variational inference such as finding a surrogate distribution $q_{\phi}(\theta)$ for the true posterior. \citet{GalG16} have shown that we can use the Dropout distribution \cite{SrivastavaHKSS14} as the approximate posterior $q_{\phi}(\theta)$. Specifically, we can perform approximate Bayesian inference by applying Dropout during test time. Hence, the posterior can now be approximated with Monte-carlo samples as follows: 
\begin{align*}
    \hat{p}(Y_1 \succ Y_2 | \mathcal{D}^{tr}) &\approx \frac{1}{L} \sum_{l=1}^{L} \hat{p}(Y_1 \succ Y_2 | \theta_{l})
\end{align*}
where $\{\theta_{l}\}_{l=1}^{L}$ are $L$ samples from the Dropout distribution $q_{\phi}(\theta)$ (i.e. we apply Dropout $L$ times independently during testing). We now discuss two different Bayesian uncertainty measures:
{\flushleft \textbf{BALD:}} The Bayesian Active Learning by Disagreement (BALD) \cite{Houlsby2011} is defined as the mutual information between the model predictions and the model posterior. Let $p_l = \hat{p}(Y_1 \succ Y_2 | \theta_{l})$, where $\theta_l \sim q_{\phi}(\theta)$, be the evaluation metric prediction using the $l^{th}$ sample $\theta_l$ from the Dropout distribution. Also, let $\bar{p} = \frac{1}{L} \sum_{l=1}^{L} p_{l}$ be the mean prediction. As shown in \cite{GalIG17}, we can approximate the BALD measure using samples from the Dropout distribution as:
\begin{align*}
    \hat{\mathbb{I}} = \mathbb{H}(\bar{p}) - \frac{1}{L} \sum_{l=1}^{L} \mathbb{H}(p_l) 
\end{align*}
where $\mathbb{H}$ is the binary cross entropy function. The BALD uncertainty score is essentially the difference in entropy of the mean prediction $\bar{p}$ and the average entropy of the individual predictions $\{p_{l}\}_{l=1}^{L}$. Hence, the BALD uncertainty score is high when the metric's mean prediction is uncertain (high entropy) but the individual predictions are highly confident (low entropy), \textit{i.e.}, when the metric produces disagreeing predictions with high confidence. 

{\flushleft \textbf{STD:}} We also adopt the standard deviation of the preference probability taken over the posterior distribution as a measure of uncertainty:
\begin{gather*}
    \sigma = \sqrt{\mbox{Var}_{\theta \sim \hat{p}(\theta | \mathcal{D}^{tr})} (\hat{p}(Y_1 \succ Y_2 | \theta))} \label{eqn:std}
\end{gather*}
Similar to BALD, we can approximate the above measure using the empirical standard deviation of samples drawn from the dropout distribution. 

Our proposed algorithm asks for human annotations only if the uncertainty measure (BALD or STD) is above a particular threshold.
\subsection{UCB Elimination}
\label{subsec:ucb_elimination}
The key idea here is to eliminate a set of "poorly performing" NLG systems using the automatic metric and perform human evaluations with the remaining set of systems. To eliminate sub-optimal systems, we first need to quantify a performance measure for the systems. We use the Copeland score \cite{CCB} which is defined as the normalized total number of pairwise wins for a system: $C_{i} = \frac{1}{k-1} \sum_{j \ne i} \mathbbm{1}({p}_{ij} > \frac{1}{2})$.  Copeland score is the highest for the top-ranked system with a value of 1 and it is less than 1 for all other systems. To estimate the Copeland score, we first predict the pairwise preference probability between any two systems $i$ and $j$ as follows:
\begin{gather*}
    \hat{p}_{ij} = \frac{1}{N} \sum_{Y_1, Y_2 \in \mathcal{D}_{ij}} \hat{p}({Y}_{1} \succ {Y}_{2}| \theta)
\end{gather*}
where $\mathcal{D}_{ij}$ is the test dataset consisting of generated texts from systems $i$ and $j$, $N$ is the total number of test examples, $\theta$ is the learned model parameters. We can now estimate the Copeland score $
\hat{C}_{i}$ using the estimated preference $\hat{p}_{ij}$ and eliminate all systems with Copeland scores below a threshold. However, a major problem with this approach is that evaluation metrics are often inaccurate and we could wrongly eliminate the true top-ranked system without performing any human evaluations. For example, consider the example where $i^*$ is the top-ranked system with $p_{i^{*}j} > 0.51$ $,\forall j \in \mathcal{S} - i$. If several of the predicted probabilities $\hat{p}_{i^{*}j}$ are less than $0.5$, our top-ranked system $i^{*}$ will receive a low estimated Copeland score and will be incorrectly eliminated. To overcome this problem, we define an Upper Confidence Bound (UCB) on the preference probability using uncertainty estimates that we described in \ref{subsec:uncertainty_aware}. Specifically, the upper confidence bound $\hat{u}_{ij}$ is given by $\hat{u}_{ij} = \hat{p}_{ij} + \alpha \hat{\sigma}_{ij}$ where $\alpha$ is a hyperparameter that controls the size of the confidence region and $\hat{\sigma}_{ij}^{2}$ is the estimated variance given by:
\begin{gather*}
    \hat{\sigma}_{ij}^{2} =  \frac{1}{N^2}
    \sum_{Y_1, Y_2 \in 
    \mathcal{D}_{ij}} \mbox{Var}_{\theta \sim q_{\phi}(\theta)}\hat{p}(Y_1 \succ Y_2 | \theta)
\end{gather*}
where $q_{\phi}(\theta)$ is the Dropout distribution. Using the upper confidence estimates $\hat{u}_{ij}$, we now define the optimistic Copeland score for a system $i$ as $\hat{C}_{i}^{u} = \frac{1}{K-1} \sum_{j \ne i} \mathbbm{1}(\hat{u}_{ij} > \frac{1}{2})$. Here, we consider a system $i$ to beat another system $j$ ($\hat{u}_{ij} > 0.5$) if either the estimated preference is high ($\hat{p}_{ij}$ is high) or if there is an high uncertainty in the estimation ($\hat{\sigma}_{ij}$ is high). In UCB Elimination, we eliminate a system only if the optimistic Copeland score is below a threshold.
\section{Experimental Setup}
In this section, we describe the (i) NLG tasks and datasets used in our experiments, (ii) automatic evaluation metrics used in our model-based algorithms, and (iii) annotation complexity measure used for comparing dueling bandit algorithms. 
\subsection{Tasks \& Datasets}
\label{subsec:task_and_datasets}
We use a total of 13 datasets spanning 5 tasks in our experiments which are summarized in table \ref{tab:dataset_stats}. \\
\textbf{Machine Translation (MT): } We use 7 human evaluation datasets collected from the WMT news translation tasks \cite{wmt15,wmt16} \textit{viz.} fin$\rightarrow$eng, rus$\rightarrow$eng, deu$\rightarrow$eng language pairs in WMT 2015 and tur$\rightarrow$eng, ron$\rightarrow$eng, cze$\rightarrow$eng, deu$\rightarrow$eng language pairs in WMT 2016. \\
\textbf{Grammatical Error Correction (GEC):} We utilize two human evaluation datasets collected by \cite{grammarly-dataset} where the source texts are from (i) student essays (FCE), and (ii) formal articles in Wikipedia (Wiki). We also use another GEC dataset collected by \cite{conll14-judgements} from the CoNLL-2014 Shared Task \cite{conll14-task}. \\
\textbf{Data-to-Text Generation:} We use the human evaluation data from the E2E NLG Challenge \cite{E2ENLGChallenge-results}. The task here is to generate natural language utterance from dialogue acts. \\
\textbf{Paraphrase Generation:} We use human evaluations of model generated English paraphrases released with the ParaBank dataset \cite{parabank}. \\
\textbf{Summarization:} We make use of the human evaluations \cite{LearningTS} of GPT3-like transformers on the TL;DR dataset \cite{tldr-dataset}.
\begin{table}[]
\centering
\resizebox{1\linewidth}{!}{
\begin{tabular}{l|l|c|c}
\Xhline{3\arrayrulewidth}
\multicolumn{1}{c|}{\multirow{2}{*}{Task}}                                               & \multicolumn{1}{c|}{\multirow{2}{*}{Dataset}} & \multicolumn{1}{c|}{\multirow{2}{*}{\# Systems}} & \multicolumn{1}{c}{\multirow{2}{*}{\begin{tabular}[c]{@{}c@{}}\# Human\\  Annotations\end{tabular}}} \\
\multicolumn{1}{c|}{}                                                                    & \multicolumn{1}{c|}{}                         & \multicolumn{1}{c|}{}                            & \multicolumn{1}{c}{}                                                                                 \\ \hline
\multirow{7}{*}{\begin{tabular}[c]{@{}l@{}}Machine\\ Translation\end{tabular}}            & WMT15 fin$\rightarrow$eng                                 & 14                                               & 31577                                                                                                 \\
                                                                                          & WMT15 rus$\rightarrow$eng                                 & 13                                               & 44539                                                                                                 \\
                                                                                          & WMT15 deu$\rightarrow$eng                                 & 13                                               & 40535                                                                                                 \\
                                                                                          & WMT16 tur$\rightarrow$eng                                 & 9                                                & 10188                                                                                                 \\
                                                                                          & WMT16 ron$\rightarrow$eng                                 & 7                                                & 15822                                                                                                 \\
                                                                                          & WMT16 cze$\rightarrow$eng                                 & 12                                               & 125788                                                                                                \\
                                                                                          & WMT16 deu$\rightarrow$eng                                 & 10                                               & 20937                                                                                                 \\ \hline
\multirow{3}{*}{\begin{tabular}[c]{@{}l@{}}Grammatical\\ Error\\ Correction\end{tabular}} & Grammarly (FCE)                               & 7                                                & 20328                                                                                                 \\
                                                                                          & Grammarly (Wiki)                              & 7                                                & 20832                                                                                                 \\
                                                                                          & CoNLL-2014 Shared Task                        & 13                                               & 16209                                                                                                 \\ \hline
Data-to-Text                                                                              & E2E NLG Challenge                             & 16                                               & 17089                                                                                                  \\ \hline
Paraphrase                                                                                & ParaBank                                      & 28                                               & 151148                                                                                                \\ \hline
Summarization                                                                             & TLDR OpenAI                                   & 11                                               & 4809     \\\Xhline{3\arrayrulewidth}
\end{tabular}}
\caption{Description of tasks and datasets with the number of NLG systems and pairwise human annotations}
\label{tab:dataset_stats}
\end{table}
We provide further details including preprocessing steps and downloadable links in appendix \ref{appendix:datasets}. 
\begin{table*}[]
\centering
\resizebox{0.999\textwidth}{!}{
\begin{tabular}{l|c|c|c|c|c|c|c|c|c|c|c|c|c}
\Xhline{3\arrayrulewidth}  
\multicolumn{1}{l|}{\multirow{2}{*}{Algorithm}} & \multicolumn{4}{c|}{WMT 2016}                                                                                                   & \multicolumn{3}{c|}{WMT 2015}                                                              & \multicolumn{2}{c|}{Grammarly}                                 & \multicolumn{1}{c|}{\multirow{2}{*}{\begin{tabular}[c]{@{}c@{}}CoNLL\\ '14 Task\end{tabular}}} & \multicolumn{1}{c|}{\multirow{2}{*}{\begin{tabular}[c]{@{}c@{}}E2E\\ NLG\end{tabular}}} & \multicolumn{1}{c|}{\multirow{2}{*}{\begin{tabular}[c]{@{}c@{}}Para-\\ Bank\end{tabular}}} & \multicolumn{1}{c}{\multirow{2}{*}{\begin{tabular}[c]{@{}c@{}}TL;\\ DR\end{tabular}}} \\ \cline{2-10}
\multicolumn{1}{l|}{}                           & \multicolumn{1}{c|}{tur-eng} & \multicolumn{1}{c|}{ron-eng} & \multicolumn{1}{c|}{cze-eng}       & \multicolumn{1}{c|}{deu-eng} & \multicolumn{1}{c|}{fin-eng} & \multicolumn{1}{c|}{rus-eng} & \multicolumn{1}{c|}{deu-eng} & \multicolumn{1}{c|}{FCE} & \multicolumn{1}{c|}{Wiki}           & \multicolumn{1}{c|}{}                                                                          & \multicolumn{1}{c|}{}                                                                   & \multicolumn{1}{c|}{}                                                                      & \multicolumn{1}{c}{}                                                                  \\ \hline 
Uniform                                          & 19479                        & 24647                        & 10262                              & 3032                         & 2837                         & 12265                        & 17795                        & 8115                     & 34443                               & 61369                                                                                          & 65739                                                                                   & 825211                                                                                     & 5893                                                                                   \\ \hline
SAVAGE                                           & 10289                        & 18016                        & 6639                               & 2393                         & 2675                         & 12806                        & 12115                        & 5767                     & 22959                               & 39208                                                                                          & 41493                                                                                   & 255208                                                                                     & 4733                                                                                   \\
DTS                                              & 10089                        & 9214                         & 8618                               & 4654                         & 4850                         & 13317                        & 16473                        & 4355                     & 11530                               & 18199                                                                                          & 19940                                                                                   & 170467                                                                                     & 1354                                                                                   \\
CCB                                              & 7017                         & 11267                        & 5389                               & 2884                         & 4092                         & 11548                        & 10905                        & 4386                     & 10020                               & 21392                                                                                          & 16960                                                                                   & 87138                                                                                      & 2518                                                                                   \\
Knockout                                         & 3415                         & 7889                         & 4723                               & 3444                         & 5104                         & 5809                         & 5956                         & 3134                     & 3777                                & \textbf{8055}                                                                                           & 7708                                                                                    & 17418                                                                                      & 4953                                                                                   \\

RUCB                                             & 3125                         & 5697                         & 3329                               & 1636                         & \textbf{1655}                & 4536                         & 6222                         & 2732                     & 5617                                & 19024                                                                                          & 10924                                                                                   & 41149                                                                                      & 1647                                                                                   \\
RCS                                              & 2442                         & \textbf{3924}                         & 3370                               & 1537                         & 2662                         & 3867                         & 5296                         & \textbf{1816}                     & \textbf{4606}                       & 12678                                                                                          & 7263                                                                                    & 34709                                                                                      & 1903                                                                                   \\
RMED                                            & \textbf{2028}                & {5113}                & \textbf{1612}                      & \textbf{864}                 & 1707                         & \textbf{1929}                & \textbf{4047}                & {2093}            & 5647                                & 9364                                                                                           & \textbf{3753}                                                                                    & \textbf{24132}                                                                             & \textbf{1162}       \\  \Xhline{3\arrayrulewidth}  
\end{tabular}}
\caption{Annotation complexity of the top 7 best performing dueling bandit algorithms along with the uniform exploration algorithm on 13  datasets spanning 5 NLG tasks}
\label{tab:results_dueling_bandit_algorithms}
\end{table*}

\subsection{Automatic NLG Evaluation Metrics}
\label{subsec:automatic_evaluation_metric}
We can predict the comparison outcome $w$ using two approaches. First, we can use pairwise probability models with existing direct assessment metrics as discussed in section \ref{sec:pairwise_probability_model}. Alternatively, we can train evaluation metrics to directly predict the comparison outcome given pairs of generated texts and context/reference as input. We discuss both these approaches below:\\
\textbf{Direct Assessment Metrics: } We experiment with a total of 10 direct assessment metrics \textit{viz.} chrF \cite{chrf}, BLEU-4 \cite{bleu}, ROUGE-L \cite{rouge}, Embedding Average \cite{embedding_average}, Vector Extrema \cite{vectorextrema}, Greedy Matching \cite{greedymatch}, Laser \cite{laser}, BertScore \cite{bertscore}, MoverScore \cite{moverscore} and Bleurt \cite{bleurt}. We mention the implementation details in appendix \ref{appendix:direct_assesment_metrics}.\\
\textbf{Pairwise Evaluation Metrics:}
We finetune the pretrained Electra-base transformer model \cite{electra} to directly predict the comparison outcome $w$. We curate task-specific human evaluation datasets consisting of tuples of the form (context/reference, hypothesis 1, hypothesis 2, label) for finetuning. Due to space constraints, we mention details on the datasets and finetuning in appendix \ref{appendix:finetuning_datasets} and \ref{appendix:finetuning_details}. For the summarization task alone, we couldn't find any pairwise human judgment dataset sufficient for finetuning the Electra model.
\subsection{Annotation Complexity Measure}
\label{subsec:annotation_complexity_measure}
To evaluate the performance of dueling bandit algorithms, we define \textit{annotation complexity} as the minimum number of human annotations needed by an algorithm to identify the top-ranked NLG system with high confidence. Let $i^*$ be the actual top-ranked system, and $\hat{i^*}(n)$ denote the estimated winner by the algorithm after $n$ human annotations, then annotation complexity is defined as:
\begin{align*}
    \min n': \forall n \geq n', P(\hat{i^*}(n) = i^*) > 1 - \delta_{acc}
\end{align*}
where $\delta_{acc}$ is the allowable failure probability \textit{i.e.} the learner can make a mistake with at most $\delta_{acc}$ probability. To compute the annotation complexity, we run each dueling bandit algorithm with 200 different random seeds and find the minimum number of human annotations after which the algorithm correctly returns the top-ranked NLG system in at least 190/200 runs (we set $\delta_{acc} = 0.05$). 
\section{Results \& Discussion}
\label{sec:results}
We discuss the performance of dueling bandits algorithms in \ref{subsec:results_dueling_bandits}, automatic metrics in \ref{subsec:results_automatic_metrics} and our proposed model-based algorithms in \ref{subsec:results_model_based}. Lastly in \ref{subsec:num_nlg_systems}, we analyze the variation of annotation complexity with the number of NLG system. 
\subsection{Analysis of Dueling Bandit Algorithms}
\label{subsec:results_dueling_bandits}
\begin{figure}
    \begin{center}
        \scalebox{.3}{\input{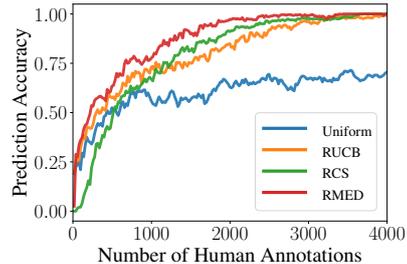}}
    \end{center}
    \caption{Top-rank prediction accuracy v/s number of human annotations used on WMT 16 tur-eng dataset}
    \label{fig:toprank_preidction_accuracy}
\end{figure}
We report the annotation complexity of the top 7 dueling bandit algorithms along with uniform exploration on 13 datasets in table \ref{tab:results_dueling_bandit_algorithms}. We observe that the annotation complexity of uniform exploration is consistently high across all 13 datasets. In particular, the required human annotations become prohibitively expensive when the number of NLG systems is high, \textit{e.g.} E2E NLG (16 systems) and ParaBank (28 systems) datasets. On the other hand, dueling bandit algorithms such as RUCB \cite{RUCB}, RCS \cite{RCS}, RMED \cite{RMED} are able to effectively identify the top-ranked system with much fewer annotations. In particular, RMED performs the best with a reduction of 80.01\% in human annotations compared to uniform exploration. We also examine an alternative approach to assess the performance of dueling bandit algorithms. Here, we fix the number of human annotations (fixed annotation budget) and compute the accuracy in predicting the top-ranked system. As we show in figure \ref{fig:toprank_preidction_accuracy}, RMED achieves the highest top-rank prediction accuracy for any given number of human annotations. We provide the complete results in appendix \ref{appendix:results_dueling_bandit_algorithms}.
\subsection{Performance of Evaluation Metrics}
\label{subsec:results_automatic_metrics}
Before we utilize automatic evaluation metrics using our proposed model-based algorithms, we analyze the effectiveness of these metrics for pairwise NLG evaluations. In table \ref{tab:results_metrics}, we report the sentence-level accuracy in predicting the comparison outcome $w$ using direct assessment metrics with the Linear probability model (as discussed in section \ref{sec:pairwise_probability_model}) along with our trained Electra metric. Across the tasks, we observe that metrics that utilize contextualized word embeddings, such as BertScore, perform much better than $n$-gram and static word embedding-based metrics. In MT, we observe that Bleurt, specifically finetuned on WMT human judgment data, performs the best. In Data-to-Text and Paraphrase generation, our trained Electra metric finetuned on task-specific data significantly outperforms the existing metrics. Interestingly, on the summarization task, all the existing metrics perform much worse than random predictions \textit{i.e.} they do not add any useful value in evaluation. Hence, we exclude the TLDR dataset from our analysis on model-based algorithms. Finally, as we show in appendix \ref{appendix:results_automatic_metrics}, we observed that the performance is largely similar across all the three probability models: Linear, BTL, and BTL-logistic.
\subsection{Analysis of Model-based Algorithms}
\begin{table}[]
\centering
\resizebox{0.999\columnwidth}{!}{
\begin{tabular}{l|c|c|c|c|c|c}
\Xhline{4\arrayrulewidth} 
\multicolumn{1}{l|}{\multirow{2}{*}{Metric}} & \multicolumn{1}{c|}{\multirow{2}{*}{\begin{tabular}[c]{@{}c@{}}WMT\\ (Avg.)\end{tabular}}} & \multicolumn{1}{c|}{\multirow{2}{*}{\begin{tabular}[c]{@{}c@{}}Gramm.\\ (Avg.)\end{tabular}}} & \multicolumn{1}{c|}{\multirow{2}{*}{\begin{tabular}[c]{@{}c@{}}CoNLL\\ '14 Task\end{tabular}}} & \multicolumn{1}{c|}{\multirow{2}{*}{\begin{tabular}[c]{@{}c@{}}E2E\\ NLG\end{tabular}}} & \multicolumn{1}{c|}{\multirow{2}{*}{\begin{tabular}[c]{@{}c@{}}Para-\\ Bank\end{tabular}}} & \multicolumn{1}{c}{\multirow{2}{*}{\begin{tabular}[c]{@{}c@{}}TL;\\ DR\end{tabular}}} \\
\multicolumn{1}{l|}{}                           & \multicolumn{1}{c|}{}                                                                      & \multicolumn{1}{c|}{}                                                                         & \multicolumn{1}{c|}{}                                                                          & \multicolumn{1}{c|}{}                                                                   & \multicolumn{1}{c|}{}                                                                      & \multicolumn{1}{c}{}                                                                  \\ \hline
Chrf                                             & 62.6                                                                                       & 75.7                                                                                          & 78.4                                                                                           & 47.4                                                                                    & 66.1                                                                                       & 34.2                                                                                   \\
Bleu                                           & 41.5                                                                                       & 73.2                                                                                          & 78.9                                                                                           & 45.0                                                                                    & 63.8                                                                                       & 42.8                                                                                   \\
Rouge-L                                          & 60.7                                                                                       & 73.5                                                                                          & 78.0                                                                                           & 44.6                                                                                    & 64.3                                                                                       & 43.3                                                                                   \\ \hline
Embed. Avg.                                    & 56.5                                                                                       & 70.1                                                                                          & 76.0                                                                                           & 49.8                                                                                    & 64.9                                                                                       & 38.2                                                                                   \\
Greedy Match.                                    & 59.5                                                                                       & 68.1                                                                                          & 77.7                                                                                           & 46.5                                                                                    & 64.7                                                                                       & 43.1                                                                                   \\
Vector Extr.                                   & 59.4                                                                                       & 66.0                                                                                          & 76.3                                                                                           & 44.9                                                                                    & 63.7                                                                                       & 47.4                                                                                   \\ \hline
BertScore                                       & 65.9                                                                                       & \textbf{77.4}                                                                                          & \textbf{82.0}                                                                                  & 45.9                                                                                    & 68.1                                                                                       & 44.5                                                                                   \\
Laser                                            & 65.3                                                                                       & 75.1                                                                                          & 78.0                                                                                           & 47.2                                                                                    & 67.0                                                                                       & 35.4                                                                                   \\
MoverScore                                       & 66.1                                                                                       & 74.7                                                                                          & 80.6                                                                                           & 50.1                                                                                    & 68.0                                                                                       & 40.7                                                                                   \\ \hline
Bleurt                                           & \textbf{68.2}                                                                              & {77.1}                                                                                 & 81.5                                                                                           & 48.1                                                                                    & 67.7                                                                                       & 42.5                                                                                   \\ 
Electra (Ours)                                   & 65.7                                                                                       & 74.0                                                                                          & 81.6                                                                                           & \textbf{54.3}                                                                           & \textbf{81.7}                                                                              & -  \\ \Xhline{4\arrayrulewidth}  
                                                                                   
\end{tabular}}
\caption{Sentence-level accuracy of direct assessment metrics with linear probability model and our trained Electra metric in predicting the comparison outcome}
\label{tab:results_metrics}
\end{table}
\begin{table*}[]
\centering
\resizebox{0.999\textwidth}{!}{
\begin{tabular}{l|l|c|c|c|c|c|c|c|c|c|c|c|c}
\Xhline{4\arrayrulewidth}
\multicolumn{1}{l|}{\multirow{2}{*}{\begin{tabular}[c]{@{}l@{}}Model-based\\ Algorithm\end{tabular}}} & \multicolumn{1}{l|}{\multirow{2}{*}{\begin{tabular}[c]{@{}l@{}}Evaluation\\ Metric\end{tabular}}} & \multicolumn{4}{c|}{WMT 2016}                                                                                             & \multicolumn{3}{c|}{WMT 2015}                                                              & \multicolumn{2}{c|}{Grammarly}                       & \multicolumn{1}{c|}{\multirow{2}{*}{\begin{tabular}[c]{@{}c@{}}CoNLL\\ '14 Task\end{tabular}}} & \multicolumn{1}{c|}{\multirow{2}{*}{\begin{tabular}[c]{@{}c@{}}E2E\\ NLG\end{tabular}}} & \multicolumn{1}{c}{\multirow{2}{*}{\begin{tabular}[c]{@{}c@{}}Para-\\ Bank\end{tabular}}} \\ \cline{3-11}
\multicolumn{1}{l|}{}                                                                                 & \multicolumn{1}{l|}{}                                                                             & \multicolumn{1}{c|}{tur-eng} & \multicolumn{1}{c|}{ron-eng} & \multicolumn{1}{c|}{cze-eng} & \multicolumn{1}{c|}{deu-eng} & \multicolumn{1}{c|}{fin-eng} & \multicolumn{1}{c|}{rus-eng} & \multicolumn{1}{c|}{deu-eng} & \multicolumn{1}{c|}{FCE} & \multicolumn{1}{c|}{Wiki} & \multicolumn{1}{c|}{}                                                                          & \multicolumn{1}{c|}{}                                                                   & \multicolumn{1}{c}{}                                                                      \\ \hline
None (Model free)                                                                                                      & None                                                                                                 & 2028                         & 5113                         & 1612                         & 864                          & 1707                         & 1929                         & 4047                         & 2093                     & 5647                      & 9364                                                                                           & 3753                                                                                    & 24132                                                                                      \\ \hline
\multirow{2}{*}{Random Mixing}                                                                            & Bleurt                                                                                            & 237                          & 1222                         & 315                          & 161                          & 275                          & 304                          & 771                          & 406                      & 671                       & 9584                                                                                           & 1151                                                                                    & 15874                                                                                      \\
                                                                                                       & Electra                                                                                           & 728                          & 3213                         & 385                          & 152                          & 236                          & 512                          & 650                          & 1529                     & 237                       & 3302                                                                                           & {326}                                                                            & 1044                                                                                       \\ \hline
\multirow{2}{*}{\begin{tabular}[c]{@{}l@{}}Uncertainty-aware\\ Selection (STD)\end{tabular}}                                                                & Bleurt                                                                                            & 103                          & 1012                         & 192                 & 84                           & 204                          & 239                          & 530                          & 270                      & 185                       & 9356                                                                                           & 1291                                                                                    & 22876                                                                                      \\
                                                                                                       & Electra                                                                                           & 978                          & 7251                         & 478                          & 210                          & 388                          & 962                          & 1259                         & 477                      & 234                       & 4708                                                                                           & 199                                                                                     & 2137                                                                                       \\ \hline
\multirow{2}{*}{\begin{tabular}[c]{@{}l@{}}Uncertainty-aware\\ Selection (BALD)\end{tabular}}                                                                     & Bleurt                                                                                            & 101                 & 653                 & \textbf{136}                 & 48                  & 181                 & {162}                 & 405                          & {204}             & {128}              & {9356}                                                                                  & 1167                                                                                    & 22619                                                                                      \\
                                                                                                       & Electra                                                                                           & 737                          & 1648                         & 223                          & 114                          & 207                          & 538                          & 488                          & 281                      & 75                        & 1557                                                                                           & 67                                                                                      & \textbf{858}                                                                               \\ \hline
\multirow{2}{*}{UCB Eliminination}                                                                           & Bleurt                                                                                            & 711                          & 2684                         & 1131                         & 573                          & 419                          & 843                          & 3556                         & 967                      & 1115                      & 8382                                                                                           & 2005                                                                                    & 14098                                                                                      \\
                                                                                                       & Electra                                                                                           & 264                          & 649                          & 1131                         & 414                          & {294}                 & 1126                         & 3556                         & 3970                     & 1115                      & 2943                                                                                           & 1112                                                                                    & {9870}                                                                              \\ \hline
\multirow{2}{*}{\begin{tabular}[c]{@{}l@{}}Uncertainty\\ (BALD) + UCB Elim.\end{tabular}}             & Bleurt                                                                                            & \textbf{31}                  & \textbf{415}                 & 376                          & \textbf{25}                  & \textbf{59}                  & \textbf{82}                  & 305                          & \textbf{162}             & \textbf{39}               & 9995                                                                                           & 256                                                                                     & 4570                                                                                       \\
                                                                                                       & Electra                                                                                           & 721                          & 736                          & 144                          & 51                           & 76                           & 288                          & \textbf{280}                 & 312                      & 45                        & \textbf{782}                                                                                   & \textbf{40}                                                                             & 2247     \\ \Xhline{4\arrayrulewidth}
\end{tabular}}
\caption{Annotation complexity of model-based algorithms when used with RMED and Bleurt/Electra metric.}
\label{tab:model_based_algorithms}
\end{table*}
\label{subsec:results_model_based}
\begin{figure}
    \begin{center}
        \scalebox{.26}{\input{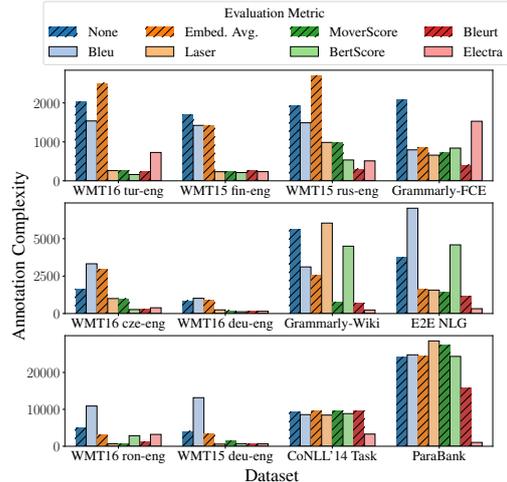}}
    \end{center}
    \caption{Annotation complexity of Random Mixing with RMED using various automatic evaluation metrics}
    \label{fig:random_mixing}
\end{figure}
We use our proposed model-based algorithms and incorporate the two best-performing evaluation metrics, \textit{viz.}, Bleurt and Electra with the best performing dueling bandit algorithm, \textit{viz.}, RMED. We compare the annotation complexity of various model-based algorithms in table \ref{tab:model_based_algorithms}. We observe that the Random Mixing algorithm with Bleurt and Electra reduces annotation complexity by 70.43\% and 73.15\%, respectively, when compared to the standard (model-free) RMED algorithm (row 1). Our Uncertainty-aware selection algorithm with the BALD measure further reduces the annotation complexity by around 37\% (compared with Random Mixing). We notice that our UCB Elimination algorithm also provides significant improvements over standard RMED. Since UCB Elimination is complementary to Uncertainty-aware selection, we apply both these algorithms together and observe the lowest annotation complexity with a reduction of 89.54\% using Electra and 84.00\% using Bleurt over standard RMED. Lastly, in figure \ref{fig:random_mixing}, we analyze the effect of using other evaluation metrics such as BLEU, BertScore, \textit{etc.}, in Random Mixing. Interestingly, we notice that using metrics such as BLEU, which have low accuracy values, results in a higher annotation complexity than standard (model-free) RMED in some datasets. That is, we may even require a greater number of human annotations to over-compensate for the inaccurate predictions from metrics like BLEU. However, with Laser, MoverScore, and BertScore, we observe significant reductions in annotation complexity. Please refer appendix \ref{appendix:model_based_algo} for further results. 
\subsection{Effect of Number of NLG systems}
\label{subsec:num_nlg_systems}
\begin{figure}
    \begin{center}
        \scalebox{.28}{\input{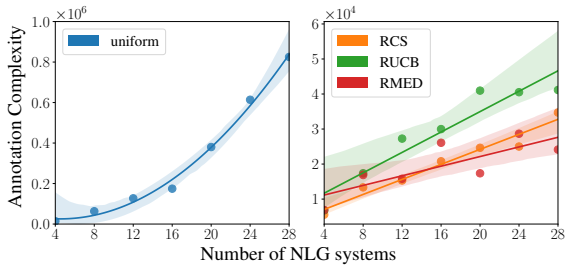}}
    \end{center}
    \caption{Annotation complexity of (model-free) uniform exploration and dueling bandit algorithms v/s the number of NLG systems on the ParaBank dataset}
    \label{fig:number_of_nlg_systems}
\end{figure}
We analyze how annotation complexity varies with the number of NLG systems. Specifically, we chose a subset of $k$ systems out of the total 28 systems in the ParaBank dataset and computed the annotation complexity among these $k$ systems. As shown in figure \ref{fig:number_of_nlg_systems}, the annotation complexity of uniform exploration grows quadratically with $k$ as it explores all system pairs equally. However, for (model-free) dueling bandit algorithms such as RMED, the annotation complexity is much lower and only varies as $O(k)$. As shown in appendix \ref{appendix:num_of_nlg_systems}, we observed similar trends with model-based algorithms.


\section{Practical Recommendations}
We summarize the key insights from this study and provide practical recommendations on efficiently identifying the top-ranked NLG system. 
\begin{enumerate}
    \item Use RMED dueling bandit algorithm to actively choose system pairs for comparison.
    \item If human evaluation datasets are available, train a metric to predict the comparison outcome directly. Otherwise, use Bleurt with any of the Linear, BTL, BTL-logistic models. 
    \item Manually annotate a few examples from the test dataset and evaluate the sentence-level accuracy of the metric. If the performance is poor (e.g., accuracy near the random baseline), do not use model-based approaches, obtain feedback only from human annotators. 
    \item If the metric is reasonably accurate, use UCB Elimination with Uncertainty-aware Selection (BALD). Tune the hyperparameters of these algorithms, if possible. Otherwise, refer appendix \ref{appendix:hyperparameter_study} for best practices developed based on analyzing the sensitivity of model-based algorithms to hyperparameters. 
    \item We can reduce the annotation time if we use multiple annotators in parallel. We observed that dueling bandit algorithms, though originally proposed for sequential annotations, are robust to asynchronous feedback from multiple annotators (Refer appendix \ref{appendix:delayed_feedback} for details).
\end{enumerate}
\section{Related Work}
\label{sec:related_work}
Several works \cite{wmt14, wmt15, Sakaguchi14, Sakaguchi16} in Machine translation and Grammatical Error Correction adopt the TrueSkill algorithm \cite{trueskill}, originally used for ranking Xbox gamers, to efficiently rank NLG systems from pairwise annotations. A recent work \cite{DurmeS18} proposes an online algorithm to rank NLG systems when we receive pairwise preference feedback in the form of a continuous scalar with bounded support. The key difference in our work is that we focus on the problem of identifying the top-rank system instead of ranking all the systems. Experimental study of dueling bandit algorithms have been limited to synthetic simulations in a few works \cite{BTM, SAVAGE}. Most others \cite{RUCB, RCS, RMED, CCB, DTS} focus on  information retrieval applications that involve evaluating search retrieval algorithms \cite{RadlinskiKJ08}. To the best of our knowledge, ours is the first work to extensively study the effectiveness of dueling bandit algorithms for NLG evaluation.

\section{Conclusion \& Future work}
In this work, we focused on the problem of identifying the top-ranked NLG system with few pairwise annotations. We formulated this problem in an Active Evaluation framework and showed that dueling bandit algorithms can reduce the number of human annotations by 80\%. We then proposed model-based algorithms to combine automatic metrics with human evaluations and showed that human annotations can be reduced further by 89\%; thereby requiring only a few hundred human annotations to identify the top-ranked system. In future work, we would like to extend our analysis to the general problem of finding the top-k ranked systems. 
\label{sec:conclusion}

\section*{Discussion on Ethics \& Broader Impact}
Evaluating Natural Language Generation (NLG) models accurately and reliably with few human annotations is an important aspect of NLG research and its real-world applications. Our work shows that we can significantly reduce the number of human annotations required to find the top-ranked NLG system with high confidence. We envision that our work will benefit a wide range of applications such as translation systems, grammatical checkers, etc., where practitioners can find the best NLG model among a set of candidates more accurately and with fewer human annotations. Despite these improvements, there are still several challenges towards reliable NLG evaluation. For example, our model-based approaches, which use automatic metrics, may be subject to biases and other undesirable mistakes, depending on the metric and how they are trained in practice. Our approach may be used to evaluate models that generate fake news, toxic content, or other harmful applications, even though it is not specifically designed for such cases. 

\section*{Acknowledgments}
We thank the Department of Computer Science and Engineering, IIT Madras, and the Robert Bosch Center for Data Science and Artificial Intelligence, IIT Madras (RBC-DSAI), for providing us resources required to carry out this research. We also wish to thank Google for providing access to TPUs through the TFRC program. We thank the anonymous reviewers for their constructive feedback in enhancing the work.

\bibliography{anthology,main}
\bibliographystyle{acl_natbib}
\newpage
\appendix
\begin{table*}[h]
\centering
\resizebox{0.8\textwidth}{!}{
\begin{tabular}{l|l|c|c|c|c}
\Xhline{3\arrayrulewidth}
\multicolumn{1}{l|}{\multirow{2}{*}{Task}}                                                & \multicolumn{1}{c|}{\multirow{2}{*}{Dataset}} & \multicolumn{1}{c|}{\multirow{2}{*}{\# Systems}} & \multicolumn{1}{c|}{\multirow{2}{*}{\begin{tabular}[c]{@{}c@{}}\# Human\\  Annotations\end{tabular}}} & \multicolumn{1}{c|}{\multirow{2}{*}{\begin{tabular}[c]{@{}c@{}}Label Distrib. \\ (0-0.5-1)\end{tabular}}} & \multirow{2}{*}{\begin{tabular}[c]{@{}c@{}}Downloadable\\  Link\end{tabular}} \\
\multicolumn{1}{l|}{}                                                                     & \multicolumn{1}{c|}{}                         & \multicolumn{1}{c|}{}                            & \multicolumn{1}{c|}{}                                                                                 & \multicolumn{1}{c|}{}                                                                                     &                                                                               \\ \hline
\multirow{7}{*}{\begin{tabular}[c]{@{}l@{}}Machine\\ Translation\end{tabular}}            & WMT15 fin-eng                                 & 14                                               & 31577                                                                                                 & 37\%-26\%-37\%                                                                                            & \multirow{3}{*}{\href{http://www.statmt.org/wmt15/translation-judgements.zip}{Click here}}                                                   \\
                                                                                          & WMT15 rus-eng                                 & 13                                               & 44539                                                                                                 & 36\%-27\%-37\%                                                                                            &                                                                               \\
                                                                                          & WMT15 deu-eng                                 & 13                                               & 40535                                                                                                 & 32\%-36\%-32\%                                                                                            &                                                                               \\ \cline{2-6}
                                                                                          & WMT16 tur-eng                                 & 9                                                & 10188                                                                                                 & 28\%-44\%-28\%                                                                                            & \multirow{4}{*}{\href{http://data.statmt.org/wmt16/translation-task/wmt16-translation-judgements.zip}{Click here}}                                                   \\
                                                                                          & WMT16 ron-eng                                 & 7                                                & 15822                                                                                                 & 38\%-24\%-38\%                                                                                            &                                                                               \\
                                                                                          & WMT16 cze-eng                                 & 12                                               & 125788                                                                                                & 38\%-25\%-37\%                                                                                            &                                                                               \\
                                                                                          & WMT16 deu-eng                                 & 10                                               & 20937                                                                                                 & 37\%-26\%-37\%                                                                                            &                                                                               \\ \hline
\multirow{3}{*}{\begin{tabular}[c]{@{}l@{}}Grammatical\\ Error\\ Correction\end{tabular}} & Grammarly (FCE)                               & 7                                                & 20328                                                                                                   & 29\%-40\%-31\%                                                                                            & \multirow{2}{*}{\href{https://github.com/grammarly/GMEG/tree/master/data/test}{Click here}}                                                   \\
                                                                                          & Grammarly (Wiki)                              & 7                                                & 20832                                                                                                   & 29\%-40\%-31\%                                                                                            &                                                                               \\ \cline{2-6}
                                                                                          & CoNLL-2014 Shared Task                        & 13                                               & 16209                                                                                                 & 23\%-52\%-25\%                                                                                            & \href{https://github.com/cnap/gec-ranking/tree/master/data}{Click here}                                                                    \\ \hline
\begin{tabular}[c]{@{}l@{}}Data-to-Text\\ Generation\end{tabular}                         & E2E NLG Challenge                             & 16                                               & 17089                                                                                                  & 24\%-50\%-26\%                                                                                            & \href{https://github.com/tuetschek/e2e-eval/releases/download/v1.0.0/e2e-eval.zip}{Click here}                                                                    \\ \hline
\begin{tabular}[c]{@{}l@{}}Paraphrase\\ Generation\end{tabular}                           & ParaBank                                      & 28                                               & 151148                                                                                                & 44\%-2\%-54\%                                                                                             & \href{https://github.com/decompositional-semantics-initiative/ParaBank-Eval-Data}{Click here}                                                                    \\ \hline
Summarization                                                                             & TLDR OpenAI                                   & 11                                               & 4809                                                                                                 & 49\%-0\%-51\%                                                                                             & \href{https://github.com/openai/summarize-from-feedback}{Click here} \\\Xhline{3\arrayrulewidth}
\end{tabular}}
\caption{Description of tasks and datasets with the number of NLG systems, number of pairwise human annotations, label distribution and the downloadable links to the datasets before preprocessing}
\label{tab:dataset_stats_appendix}
\end{table*}
\section{Further Details on Experiments}
\label{appendix:experiment_details}

\subsection{Tasks \& Datasets}
\label{appendix:datasets}
In table \ref{tab:dataset_stats_appendix}, we report the dataset statistics along with links to download the original datasets. We now discuss the preprocessing steps: \\
\textbf{Machine Translation:} In WMT 2015 and 2016 tasks, human annotators were asked to rank five system outputs (translated sentences) relative to each other. As recommended by the organizers \cite{wmt14}, we convert each of these rankings into ${5 \choose 2}$ pairwise comparisons of systems. \\
\textbf{Grammatical Error Correction:} The Grammarly evaluation datasets follow the RankME \cite{RankME} annotation style where annotators were shown 8 outputs side by side for each input and were asked to provide a numerical score to each of them. We discarded one of the outputs out of the 8, which was human crafted, and used the remaining 7 model-generated outputs. We then convert these 7 scores into ${7 \choose 2}$ pairwise comparisons of systems. Human evaluations of the CoNLL-2014 Shared Task followed the same process as WMT 2015. Hence, we follow the same preprocessing steps as WMT.\\
\textbf{Data-to-Text Generation:} The E2E NLG Challenge also follows the RankME annotation format. We follow the same preprocessing steps as the Grammarly datasets. Out of the total 21 systems, we held out 5 systems to train the Electra model and use the remaining 16 systems. \\
\textbf{Paraphrase Generation:} For ParaBank, we follow the same preprocessing steps as the Grammarly datasets. Out of the total 35 systems, we held out of 7 systems and only used the remaining 28 systems.\\
\textbf{Summarization:} We select 11 systems that have human annotations between each pair of them. These systems are GPT3-like models with varying model sizes (3B, 6B, 12B) and training strategies. We do not perform any additional preprocessing here.

\subsection{Direct Assessment Metrics: Implementation Details}
\label{appendix:direct_assesment_metrics}
We use the nlg-eval library\footnote{https://github.com/Maluuba/nlg-eval} for the implementation of BLEU-4, ROUGE-L, Embedding Average, Vector Extrema, and Greedy Matching. For chrF, Laser and BertScore, we use the implementations from the VizSeq library \footnote{https://github.com/facebookresearch/vizseq}.  We use the official implementation released by the original authors for MoverScore and Bleurt. Among these metrics, Bleurt is the only trainable metric. We use the publicly released Bleurt-base checkpoint trained on WMT direct judgments data. As described in section \ref{subsec:uncertainty_aware}, we apply Dropout to the Bleurt model during test time to estimate prediction uncertainty.
\subsection{Finetuning Datasets}
\label{appendix:finetuning_datasets}
Here, we describe the task-specific datasets used for finetuning the Electra model (pairwise evaluation metric described in section \ref{subsec:automatic_evaluation_metric}). For MT, we used human evaluations of WMT 2013 and 2014, consisting of a total of 650k examples. For GEC, we curated a training dataset of 180k pairs of texts and human preference using data released by \cite{gec_data} and the development set released by \cite{grammarly-dataset}. We utilize 11k examples from 5 held-out systems in the E2E NLG Challenge (apart from the 16 systems used for evaluations) for Data-to-Text generation. Lastly, we use a dataset of 180k examples from 7 held-out systems in the ParaBank dataset for paraphrase generation.  We use $90\%-10\%$ split for splitting the dataset into train and validation sets. Note that these datasets do not have any overlap with the datasets used for evaluating dueling bandit algorithms.

\subsection{Finetuning Details}
\label{appendix:finetuning_details}
We use the pretrained Electra-base model \cite{electra} with 110M parameters (12 layers and 12 attention heads) as our base model. We finetune the model using ADAM optimizer with $\beta_1 = 0.9$ and $\beta_2 = 0.99$. We use a linear learning rate decay with a maximum learning rate of 1e-5 and warm-up for 10\% of training. We use a batch size of 128 and finetune for four epochs. We finetune all the models on Google Cloud TPU v3-8. To estimate prediction, we apply Dropout to the Electra model during test time as described in \ref{subsec:uncertainty_aware}.

\section{Summary of Dueling Bandit Algorithms}\label{appendix:summary_dueling_bandits}
We now provide an overview of various dueling bandit algorithms in the literature. We first introduce a few additional notations and terminologies in \ref{appendix:bandits_notations}. Later in \ref{appendix:bandits_assumptions}, we describe the various structural assumptions made by different dueling bandit algorithms. Finally, in \ref{appendix:bandits_algorithms}, we summarize 13 dueling bandit algorithms that we analyze in this work.
\subsection{Notations and Terminologies}
\label{appendix:bandits_notations}
Let $\Delta_{ij} = p_{ij} - \frac{1}{2}$ where $p_{ij}$ is the preference probability of system $i$ over $j$, as defined in section \ref{subsec:top_rank_system}. We call a system as the Copeland winner if it beats more number of systems than any other system. Mathematically, a Copeland winner $i^*$ is defined as $i^* = \arg \max_{i} \sum_{j=1}^{k} \mathbbm{1}(\Delta_{ij} > 0)$. A special case of the Copeland winner is the Condorcet winner, which is the system that beats all other systems. In all our NLG tasks and datasets, we observed that this special case holds true \textit{i.e.} there exists a system that beats all other $k-1$ systems, and we define it as the top-ranked system. Nevertheless, we mention these two definitions to distinguish algorithms that work for the general Copeland winner, even if the Condorcet winner does not exist. 
\subsection{Assumptions}\label{appendix:bandits_assumptions}
All the dueling bandit algorithms that we analyze in this work assume a stochastic feedback setup in which the feedback is generated according to an underlying (unknown) stationary probabilistic process. Specifically, in our Active Evaluation framework, this is equivalent to assuming that the annotator preference is stationary over time and is given by some fixed distribution $p_a(w |{Y}^{(t)}_1, {Y}^{(t)}_2)$. Further, many dueling bandit algorithms make various assumptions on the true pairwise preferences and exploit these assumptions to derive theoretical guarantees \cite{dueling_bandits_survey}. In table \ref{tab:assumptions}, we describe the various commonly used assumptions by dueling bandit algorithms. For example, the stochastic triangle inequality assumption (STI), described in row 4 of table \ref{tab:assumptions}, assumes that the true preference probabilities between systems obey the triangle inequality. We note here that one cannot verify the validity of these assumptions apriori since we do not have access to the true preferences. 
\subsection{Algorithms}\label{appendix:bandits_algorithms}
In table \ref{tab:dueling_bandit_algorithms}, we describe the various dueling bandit algorithms along with the assumptions (used to provide theoretical guarantees) and the target winner. We summarize these algorithms below:\\
{\flushleft \textbf{IF:}} Interleaved Filtering (IF) \cite{Yue2009TheKD}  algorithm consists of a sequential elimination strategy where a currently selected system $s_i$ is compared against the rest of the active systems (not yet eliminated). If the system $s_j$ beats a system $s_i$ with high confidence, then $s_i$ is eliminated, and $s_j$ is compared against all other active systems. Similarly, if the system $s_i$ beats $s_j$ with high confidence, then $s_j$ is eliminated, and $s_i$ is continued to be compared against the remaining active systems. Under the assumptions of TO, SST, and STI, the authors provide theoretical guarantees for the expected regret achieved by IF.
{\flushleft \textbf{BTM:}} Beat The Mean (BTM) \cite{BTM}, similar to IF, is an elimination-based algorithm that selects the system $s_i$ with the fewest comparisons and compares it with a randomly chosen system from the set of active systems. Based on the comparison outcome, a score and confidence interval are assigned to the system $s_i$. BTM eliminates a system as soon as there is another system with a significantly higher score. 
{\flushleft \textbf{Knockout, Seq Elim, Single Elim:}} Knockout \cite{knockout}, Sequential Elimination \cite{sequential_elimination}, Single Elimination \cite{single_elimination} are all algorithms that proceed in a  knockout tournament fashion where the systems are randomly paired, and the winner in each duel will play the next round (losers are knocked out) until the overall winner is determined. During a duel, the algorithm repeatedly compares the two systems to reliably determine the winner. The key difference between the three algorithms is the assumptions they use and how they determine the number of comparisons required to identify the winning system in a duel with high probability. 
{\flushleft \textbf{Plackett Luce:}} Plackett Luce Condorcet winner identification algorithm \cite{Plackettluce} assumes that the true rank distribution follows the Placket-Luce model \cite{Plackett1975TheAO}. The algorithm is based on a budgeted version of QuickSort. The authors show that it achieves a worst-time annotation complexity of the order $k \log k$ under the Placket-Luce assumption.
{\flushleft \textbf{RUCB:}} Relative Upper Confidence Bound (RUCB) \cite{RUCB} is an adaptation of the well-known UCB algorithm \cite{UCB} to the dueling bandit setup. Similar to UCB, RUCB selects the first system $s^{(1)}_{t}$ based on "optimistic" estimates of the pairwise preference probabilities \textit{i.e.} based on an upper confidence bound of preference probabilities. The second system $s^{(2)}_{t}$ is chosen to be the one that is most likely to beat $s^{(1)}_{t}$. 
{\flushleft \textbf{RCS:}} Relative Confidence Sampling (RCS) \cite{RCS} follows a Bayesian approach by maintaining a posterior distribution over the preference probabilities. At each time step $t$, the algorithm samples preference probabilities from the posterior and simulates a round-robin tournament among the systems to determine the Condorcet winner. The estimated Condorcet winner is chosen as the first system $s^{(1)}_{t}$ and second system $s^{(2)}_{t}$ is chosen such that it has the best chance of beating $s^{(1)}_{t}$.
\begin{table}[]
\resizebox{1.0\columnwidth}{!}{
\begin{tabular}{l|l}
\Xhline{3\arrayrulewidth}
Assumption Name                                                                            & Condition                                                                                                                                                                                                                                                                                            \\ \hline
Total Order (TO)                                                                & \begin{tabular}[c]{@{}l@{}}$\exists$ a total order $\succ$ over $\mathcal{S}$: \\ $i \succ j \iff \Delta_{ij} > 0$\end{tabular} \\ \hline
\begin{tabular}[c]{@{}l@{}}Strong stochastic \\ transitivity (SST)\end{tabular} & \begin{tabular}[c]{@{}l@{}}$\Delta_{ij}>0, \Delta_{jk}>0 \implies$\\ $\Delta_{ik} \geq \max (\Delta_{ij}, \Delta_{jk})$\end{tabular} \\  \hline
\begin{tabular}[c]{@{}l@{}}Relaxed stochastic\\ transitivity (RST)\end{tabular} & \begin{tabular}[c]{@{}l@{}}$ \exists \gamma \geq 1$: $\Delta_{ij}>0, \Delta_{jk}>0 $\\ $\implies \gamma \Delta_{ik} \geq \max (\Delta_{ij}, \Delta_{jk})$\end{tabular} \\  \hline
\begin{tabular}[c]{@{}l@{}}Stochastic triangle \\ inequality (STI)\end{tabular} & \begin{tabular}[c]{@{}l@{}}$\Delta_{ij}>0, \Delta_{jk}>0 \implies$\\  $\Delta_{ik} \leq \Delta_{ij} + \Delta_{jk}$\end{tabular} \\   \hline
Condorcet winner (CW)                                                           & $\exists i^*$: $\Delta_{i^*,j} > 0, \forall j \in \mathcal{S} - i^*$                                                                                                                                                                                                                                 \\  \hline
PL model                                                                        & \begin{tabular}[c]{@{}l@{}}The underlying rank distribution \\ follows the Plackett-Luce (PL) \\ model \cite{Plackett1975TheAO, Luce1979IndividualCB}\end{tabular}      \\  \Xhline{3\arrayrulewidth}
\end{tabular}}
\caption{Various assumptions made by dueling bandit algorithms in the literature}
\label{tab:assumptions}
\end{table}
{\flushleft \textbf{RMED:}}  Relative Minimum Empirical Divergence1 (RMED) algorithm \cite{RMED} maintains an empirical estimate of the “likelihood” that a system is the Condorcet winner. It then uses this estimate to sample the first system $s^{(1)}_{t}$ and then selects the second system $s^{(2)}_{t}$ that is most likely to beat $s^{(1)}_{t}$. 
{\flushleft \textbf{SAVAGE:}} Sensitivity Analysis of VAriables for Generic Exploration (SAVAGE) \cite{SAVAGE} is a generic algorithm that can be adopted for various ranking problems such as Copeland winner identification. SAVAGE (Copeland) algorithm, at each time step, randomly samples a pair of systems from the set of active system pairs (not yet eliminated) and updates the preference estimates. A system pairs $(s_i, s_j)$ is eliminated if either (i) the result of comparison between $s_i$ and $s_j$ is already known with high probability, or (ii) there exists some system $s_k$ where the estimated Copeland score of $s_k$ is significantly higher than $s_i$ or $s_j$. 
\begin{table}[]
\resizebox{1.0\columnwidth}{!}{
\begin{tabular}{lcc}
\Xhline{3\arrayrulewidth}
Algorithm             & Assumptions & Target\\ \hline
IF \cite{Yue2009TheKD}                   & TO+SST+STI & Condorcet                                                         \\
BTM \cite{BTM}                  & TO+RST+STI & Condorcet                                                          \\
Seq-Elim. \cite{sequential_elimination}
&  SST  & Condorcet                                               \\
Plackett Luce \cite{Plackettluce}     &   PL model         & Condorcet                                               \\
Knockout \cite{knockout}  &     SST+STI       & Condorcet                                               \\
Single Elim.\cite{single_elimination}   &  TO          & Condorcet                                               \\ \hline
RUCB \cite{RUCB}                  & CW         & Condorcet                                                          \\
RCS \cite{RCS}                  & CW         & Condorcet                                                          \\
RMED \cite{RMED}                & CW         & Condorcet \\ \hline 
SAVAGE \cite{SAVAGE}               & -         & Copeland                                               \\
CCB \cite{CCB}                  & -    & Copeland                                                          \\
DTS \cite{DTS}                  & -    & Copeland                                                          \\
DTS++ \cite{DTS}                & -    & Copeland                                                         \\ \Xhline{3\arrayrulewidth}
\end{tabular}}
\caption{Summary of dueling bandits algorithms in the literature along with their theoretical assumptions and the target winner of the learner}
\label{tab:dueling_bandit_algorithms}
\end{table}
{\flushleft \textbf{CCB:}} Copeland Confidence Bound (CCB) \cite{CCB} is similar to the RUCB algorithm but is designed to identify the Copeland Winner (a generalization of the Condorcet winner). The CCB algorithm maintains optimistic preference estimates and uses them to choose the first system $s^{(1)}_t$ and then selects the second system $s^{(2)}_t$ that is likely to discredit the hypothesis that $s^{(1)}_t$ is indeed the Copeland winner. The algorithm successively removes all other systems that are highly unlikely to be a Copeland winner. 
{\flushleft \textbf{DTS, DTS++:}}  The Double Thompson Sampling (DTS) algorithm \cite{DTS} maintains a posterior distribution over the pairwise preference matrix, and selects the system pairs $s^{(1)}_t, s^{(2)}_t$ based on two independent samples from the posterior distribution. The algorithm updates the posterior distributions based on the comparison outcome and eliminates systems that are unlikely to be the Copeland winner. DTS++ is an improvement proposed by the authors, which differs from DTS in the way the algorithm breaks ties. Both have the same theoretical guarantees, but DTS++ has been empirically shown to achieve better performance (in terms of regret minimization).  
\section{Hyperparameters Details}
We discuss the details of the hyperparameters and the tuning procedure used for dueling bandit algorithm in \ref{appendix:hyperparam_bandits}, pairwise probability models in \ref{appendix:pairwise_probability_models} and our model-based algorithm in \ref{appendix:hyperparam_model_based}. In all three cases, we use the validation split of the finetuning datasets described in \ref{appendix:finetuning_datasets} as our validation dataset. For example, the validation split of the finetuning datasets for MT consists of 10\% of the WMT 2013 and 2014 datasets. We use this dataset to tune the hyperparameters for  WMT 2015 and 2016 datasets. 
\subsection{Dueling Bandit Algorithms}
\label{appendix:hyperparam_bandits}
For all algorithms other than Knockout and Single Elimination, we use the hyperparameters recommended by the original authors for all the datasets. For example, in the RMED algorithm, described in algorithm 1 of \cite{RMED}, we use $f(K) = 0.3K^{1.01}$ as suggested by the authors. For the RCS algorithm, described in algorithm 1 of \cite{RCS}, we use $\alpha$ (exploratory constant) $= 0.501$. For RUCB (algorithm 1 of \cite{RUCB}), we use $\alpha = 0.51$. Similarly, for all algorithms other than Knockout and Single Elimination, we use the recommended hyperparameters mentioned in the original paper. For knockout and Single Elimination, we found that the performance was very sensitive to the hyperparameters. For these two algorithms, we manually tuned the hyperparameters on the validation set. In Knockout, algorithm 3 of \cite{knockout}, we use $\epsilon = 0.2, \delta = 0.05, \gamma = 1.0$ for WMT'16 ron-eng and TLDR OpenAI datasets. We use $\epsilon = 0.2, \delta = 0.05, \gamma = 0.6$ for ParaBank and Grammarly-Wiki datasets and  $\epsilon = 0.2, \delta = 0.09, \gamma = 0.6$ for all other datasets. In Single Elimination, we use $m$ (number of pairwise comparisons per duel) $=1000$ for WMT'16 ron-eng, E2E NLG, Grammarly-FCE, $m=1500$ for CoNLL'14 shared task and $m=500$ for all other datasets.

\subsection{Pairwise Probability Models}
\label{appendix:pairwise_probability_models}
Let $\Tilde{f}(Y)$ be the unnormalized score given an automatic evaluation metric for an hypothesis $Y$. We preprocess the score $\Tilde{f}(Y)$ to obtain $f(Y)$ to ensure that the pairwise probability scores is always a valid \textit{i.e.} lies between 0 and 1. To preprocess the scores, we use the validation dataset consisting of tuples of the form $\{Y_{1}^{(i)}, Y_{2}^{(i)}, w^{(i)}\}_{i=1}^{N}$ where $Y_{1}^{(i)}$, $Y_{2}^{(i)}$ represent the $i$th generated texts and $w^{(i)}$ is the corresponding comparison outcome provided by human annotators.\\

\textbf{Linear: } Let $\Delta_i = |\Tilde{f}(Y_{1}^{(i)}) - \Tilde{f}(Y_{2}^{(i)})|$ and $\Delta = \max_{i} \Delta_i$. We divide the unormalized $\Tilde{f}(Y)$ scores by $2\Delta$ \textit{i.e.}
$$ f(Y) = \frac{\Tilde{f}(Y)}{2\Delta}$$.  

\textbf{BTL:} Let $f^{m}_i = \max \{\Tilde{f}(Y^{(i)}_1), \Tilde{f}(Y^{(i)}_2)\}$, $f^{m} = \max_i f^{m}_i$. We now subtract the scores by $f^{m}$ to ensure that the scores are non-negative \textit{i.e.}
$$ f(Y) = \Tilde{f}(Y) - f^{m}$$ 

\textbf{BTL-Logistic:} BTL-Logistic model always provides a score between 0 and 1. However, we found that dividing the scores by a temperature co-efficient $\gamma$ can provide better results \textit{i.e.}
$$ f(Y) = \frac{\Tilde{f}(Y)}{\gamma}$$
We tune $\gamma$ using grid search between 0.005 and 1 on the validation set to minimize the cross-entropy loss between the preference probabilities $\hat{p}({Y}_{1} \succ {Y}_{2})$ and the human labels $w$. \\
 {\flushleft \textbf{Thresholds:}} As described in section \ref{sec:pairwise_probability_model}, we threshold the preference probabilities $\hat{p}({Y}_{1} \succ {Y}_{2})$ at two thresholds $\tau_1$ and $\tau_2$ to obtain the predicted comparison outcome $\hat{w}$. We perform a grid search by varying $\tau_1$ from 0.4 to 0.5 and $\tau_2$ from 0.5 to 0.6 with a step size of 0.001. We choose the optimal thresholds that maximize the prediction accuracy on the validation dataset. 

\subsection{Model-based Algorithms}
\label{appendix:hyperparam_model_based}
We manually tune the hyperparameters in our model-based algorithms on the validation dataset. For clarity, we first describe the hyperparameters in the different model-based algorithms. In Random Mixing, we need to choose the mixing probability $p_m$ hyperparameter. In Uncertainty-aware Selection (BALD), we need to choose a threshold value $\tau_{BALD}$ for the BALD score at which we decide to ask for human annotations. For UCB elimination, we should choose a threshold $\tau_{cop}$ for optimistic Copeland scores and the $\alpha$ hyperparameter, which controls the size of the confidence region. In table \ref{tab:hyperparams_model_based_algorithms_electra} and \ref{tab:hyperparams_model_based_algorithms_bleurt}, we report the tuned hyperparameter values when using Electra and Bleurt (with the Linear probability model) as the evaluation model. Another hyperparameter is the number of Monte-Carlo samples $L$ to obtain from the Dropout distribution as discussed in section \ref{subsec:uncertainty_aware}. We set $L=20$, \textit{i.e.} we independently apply dropout 20 times for each test predictions. 
\begin{table}[]
\centering
\resizebox{0.999\linewidth}{!}{
\begin{tabular}{l|c|c|c|c}
\Xhline{3\arrayrulewidth}  
\multicolumn{1}{c|}{\multirow{2}{*}{Dataset}}                     & \multicolumn{1}{c|}{Rand. Mix.} & \multicolumn{1}{c|}{\begin{tabular}[c]{@{}c@{}}Uncertainty\\(BALD)\end{tabular}} & \multicolumn{2}{c}{UCB-Elim.}                     \\ \cline{2-5} 
\multicolumn{1}{c|}{}                                             & \multicolumn{1}{c|}{$p_m$}          & \multicolumn{1}{c|}{$\tau_{BALD}$}                                                                       & \multicolumn{1}{c|}{$\alpha$} & \multicolumn{1}{c}{$\tau_{cop}$} \\ \hline
\begin{tabular}[c]{@{}l@{}}WMT \\ (all 7 datasets)\end{tabular}    & 0.8                                & 0.025                                                                                              & 0.5                        & 0.8                         \\ \hline
\begin{tabular}[c]{@{}l@{}}Grammarly \\ (FCE \& Wiki)\end{tabular} & 0.8                                & 0.07                                                                                               & 0.5                        & 0.8                         \\ \hline
CoNLL'14                                                           & 0.8                                & 0.07                                                                                               & 0.5                        & 0.8                         \\ \hline
E2E NLG                                                            & 0.9                                & 0.035                                                                                              & 0.5                        & 0.8                         \\ \hline
ParaBank                                                           & 0.95                               & 0.15                                                                                              & 0.5                        & 0.8                         \\ \Xhline{3\arrayrulewidth}  
\end{tabular}}
\caption{Tuned Hyperparameters of Model-based algorithms when used with the Electra Metric}
\label{tab:hyperparams_model_based_algorithms_electra}
\end{table}
\section{Effect of Hyperparameters in Model-based Algorithms}
\label{appendix:hyperparameter_study}
\subsection{Sensitivity to Hyperparameters}
\begin{figure}
    \begin{center}
        \scalebox{.22}{\input{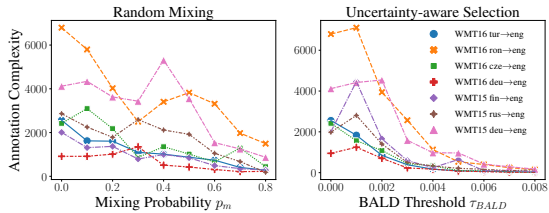}}
    \end{center}
    \caption{Variation in annotation complexity with Mixing probability in Random Mixing with Bleurt on the left and with BALD threshold in Uncertainty-aware Selection (BALD) with Bleurt on the right}
    \label{fig:hyperparameter_variation}
\end{figure}

\begin{figure}
    \centering
    \includegraphics[width=1.0\linewidth]{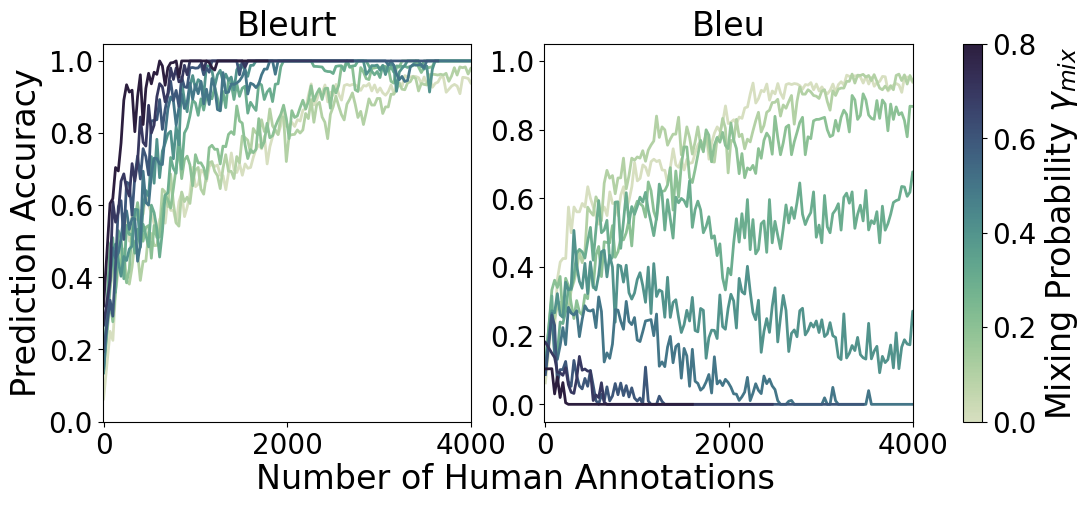}
    \caption{Prediction accuracy v/s number of human annotations collected for Random Mixing with Bluert and BLEU for different mixing probability $p_{m}$ on the WMT 15 deu-eng dataset}
    \label{fig:bleurt_vs_bleu}
\end{figure}

\begin{figure}
    \begin{center}
        \scalebox{.22}{\input{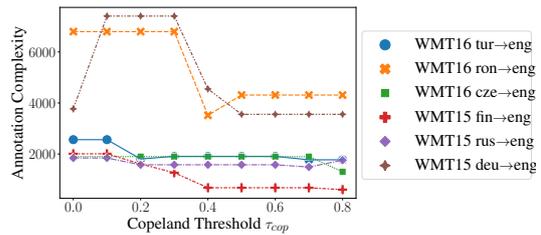}}
    \end{center}
    \caption{Annotation complexity of UCB Elimination with Bleurt v/s the Copland threshold for $\alpha=0.6$}
    \label{fig:hyperparameter_variation_ucbelim}
\end{figure}

We study how hyperparameters in our proposed model-based algorithms affect annotation complexity. Recall that in Random Mixing, the mixing probability $p_m$ controls the ratio of real and model generated feedback given to the learner. In Uncertainty-aware Selection (BALD), we obtain human annotations when the BALD score is above a threshold $\tau_{BALD}$. Here, as well $\tau_{BALD}$ implicitly controls the fraction of real and predicted feedback. In figure \ref{fig:hyperparameter_variation}, we show the effect of $p_{m}$ in Random Mixing with Bleurt and $\tau_{BALD}$ in Uncertainty-aware Selection with Bleurt. We observe that with increases in both the hyperparameters, the annotation complexity decreases, \textit{i.e.}, with a greater amount of feedback received from Bleurt, the number of required human annotations is lower. However, as shown in figure \ref{fig:bleurt_vs_bleu}, we observe the opposite trend when we use metrics such as BLEU, which are highly inaccurate. In these cases, we require a greater number of human annotations to compensate for the highly erroneous feedback received from the evaluation metric. Therefore, the optimal mixing probability $p_{m}$ in such cases is close to 0 \textit{i.e.} equivalent to the model-free case. For moderately accurate metrics such as Laser, we observed the optimal $p_{m}$ was close to 0.4 to 0.6. The key insight from these observations is that the higher the accuracy of the metric, the higher amount of feedback can be obtained from the metric to identify the top-ranked system. In figure \ref{fig:hyperparameter_variation_ucbelim}, we analyze how the annotation complexity of UCB Elimination with Bleurt varies with the optimistic Copeland threshold $\tau_{cop}$ hyperparameter. We fixed $\alpha$ hyperparameter to 0.6. We observed that UCB Elimination is much more robust to $\tau_{cop}$ and a general value of $\tau_{cop} = 0.8$ worked well across all datasets and metrics. 
\begin{table}[]
\centering
\resizebox{0.999\linewidth}{!}{
\begin{tabular}{l|c|c|c|c}
\Xhline{3\arrayrulewidth}  
\multicolumn{1}{c|}{\multirow{2}{*}{Dataset}}                     & \multicolumn{1}{c|}{Rand. Mix.} & \multicolumn{1}{c|}{\begin{tabular}[c]{@{}c@{}}Uncertainty\\(BALD)\end{tabular}} & \multicolumn{2}{c}{UCB-Elim.}                     \\ \cline{2-5} 
\multicolumn{1}{c|}{}                                             & \multicolumn{1}{c|}{$p_m$}          & \multicolumn{1}{c|}{$\tau_{BALD}$}                                                                       & \multicolumn{1}{c|}{$\alpha$} & \multicolumn{1}{c}{$\tau_{cop}$} \\ \hline
\begin{tabular}[c]{@{}l@{}}WMT \\ (all 7 datasets)\end{tabular}    & 0.8                                & 0.005                                                                                              & 0.5                        & 0.8                         \\ \hline
\begin{tabular}[c]{@{}l@{}}Grammarly \\ (FCE \& Wiki)\end{tabular} & 0.8                                & 0.0005                                                                                               & 0.5                        & 0.8                         \\ \hline
CoNLL'14                                                           & 0.01                                & 0.00005                                                                                               & 1                        & 0.7                         \\ \hline
E2E NLG                                                            & 0.7                                & 0.0025                                                                                              & 0.5                        & 0.8                         \\ \hline
ParaBank                                                           & 0.4                               & 0.0005                                                                                              & 0.5                        & 0.8                         \\ \Xhline{3\arrayrulewidth}  
\end{tabular}}
\caption{Tuned Hyperparameters of Model-based algorithms when used with the Bleurt Metric}
\label{tab:hyperparams_model_based_algorithms_bleurt}
\end{table}
\subsection{Best Practices in Choosing Hyperparameters}
The optimal approach to choose hyperparameters is usually to tune them on a validation set. But, at times, it may not be possible either because of computational reasons or because a human-annotated validation dataset may not be available. In such cases, we provide a few heuristics based on our previous analysis to choose hyperparameters in our model-based algorithms: 
\begin{enumerate}
    \item Choose the mixing probability $p_m$ in Random Mixing proportionately with the accuracy of the metric. For example, we observed that for metrics with sentence-level prediction accuracy greater than $70\%$, $p_m = 0.8$ tend to work well. For accuracy between $65\%$ to $70\%$, $p_m$ in the range of 0.5-0.7 worked well.  
    \item Once we choose a value of $p_m$, we can find an appropriate BALD threshold $\tau_{BALD}$ where $100\times p_m \%$ of BALD scores are above $\tau_{BALD}$ and $100\times (1-p_m) \%$ of BALD score are below $\tau_{BALD}$. Choosing the BALD threshold this way ensures that we can directly control the desired amount of model-predicted feedback given to the learner. 
    \item For UCB Elimination, we recommend using the default values of $\alpha=0.6$ and $\tau_{cop} = 0.8$, which we found to work well across tasks and metrics. 
\end{enumerate}

\section{Robustness to Delayed Feedback}
\label{appendix:delayed_feedback}
\begin{figure}
    \begin{center}
        \scalebox{.22}{\input{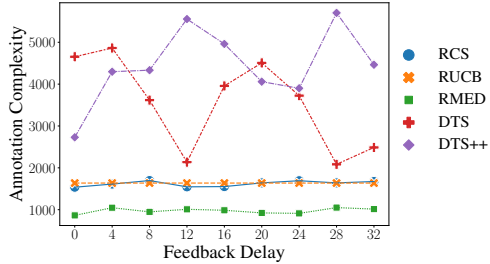}}
    \end{center}
    \caption{Annotation Complexity v/s delays in feedback on the WMT16 deu-eng dataset}
    \label{fig:feedback_delay}
\end{figure}

In some instances, human annotations are obtained from multiple crowdsourced annotators in parallel to reduce the time taken for annotations. In such cases, the learner is required to choose the system pairs $(s_{1}^{(t)}, s_{2}^{(t)})$ to give to some annotator $i$ even before we obtain the result ${w}^{(t-1)}$ of the previous comparison from some other annotator $j$. In other words, the learner may experience a delay $d > 0$ in feedback where at time $t$, the learner may only have access to the comparison history up to time $t-d-1$. As shown in figure \ref{fig:feedback_delay}, we observe that the top-performing dueling bandit algorithms tend to be robust to delays in feedback. We notice that the variation in the annotation complexity of RMED and RCS as measured by standard deviation is only 64.49 and 62.86, respectively.
\begin{figure}
    \begin{center}
        \scalebox{.2}{\input{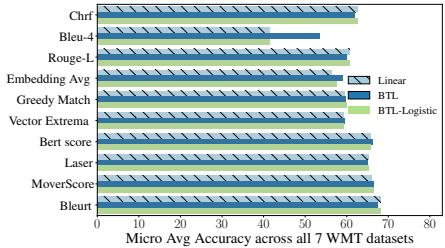}}
    \end{center}
    \caption{Sentence-level prediction accuracy of direct assessment metrics with the Linear, BTL, and BTL-Logistic models averaged across the 7 WMT datasets}
    \label{fig:probability_models}
\end{figure}
\begin{figure}
    \begin{center}
        \scalebox{.28}{\input{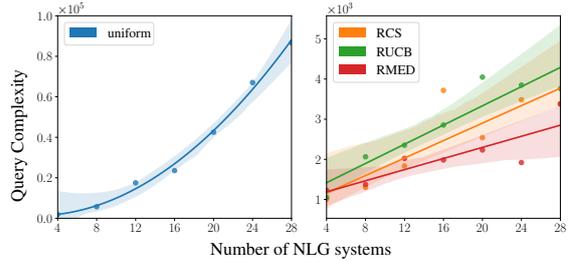}}
    \end{center}
    \caption{Annotation complexity of Random Mixing using the Electra metric with uniform exploration and dueling bandit algorithms as function of number of NLG systems on the ParaBank dataset}
    \label{fig:number_of_nlg_systems_rand_mix}
\end{figure}
\begin{table*}
\centering
\resizebox{0.999\textwidth}{!}{
\begin{tabular}{l|c|c|c|c|c|c|c|c|c|c|c|c|c}
\Xhline{3\arrayrulewidth}  
\multicolumn{1}{l|}{\multirow{2}{*}{Algorithm}} & \multicolumn{4}{c|}{WMT 2016}                                                                                                   & \multicolumn{3}{c|}{WMT 2015}                                                              & \multicolumn{2}{c|}{Grammarly}                                 & \multicolumn{1}{c|}{\multirow{2}{*}{\begin{tabular}[c]{@{}c@{}}CoNLL\\ '14 Task\end{tabular}}} & \multicolumn{1}{c|}{\multirow{2}{*}{\begin{tabular}[c]{@{}c@{}}E2E\\ NLG\end{tabular}}} & \multicolumn{1}{c|}{\multirow{2}{*}{\begin{tabular}[c]{@{}c@{}}Para-\\ Bank\end{tabular}}} & \multicolumn{1}{c}{\multirow{2}{*}{\begin{tabular}[c]{@{}c@{}}TL;\\ DR\end{tabular}}} \\ \cline{2-10}
\multicolumn{1}{l|}{}                           & \multicolumn{1}{c|}{tur-eng} & \multicolumn{1}{c|}{ron-eng} & \multicolumn{1}{c|}{cze-eng}       & \multicolumn{1}{c|}{deu-eng} & \multicolumn{1}{c|}{fin-eng} & \multicolumn{1}{c|}{rus-eng} & \multicolumn{1}{c|}{deu-eng} & \multicolumn{1}{c|}{FCE} & \multicolumn{1}{c|}{Wiki}           & \multicolumn{1}{c|}{}                                                                          & \multicolumn{1}{c|}{}                                                                   & \multicolumn{1}{c|}{}                                                                      & \multicolumn{1}{c}{}                                                                  \\ \Xhline{3\arrayrulewidth}  
Uniform                                          & 19479                        & 24647                        & 10262                              & 3032                         & 2837                         & 12265                        & 17795                        & 8115                     & 34443                               & 61369                                                                                          & 65739                                                                                   & 825211                                                                                     & 5893                                                                                   \\ \hline
IF                                               & 117762                       & 282142                       & 135718                             & 75014                        & 101380                       & 162536                       & 261300                       & 226625                   & 364304                              & 713522                                                                                         & 718492                                                                                  & 605825                                                                                     & 70071                                                                                  \\
BTM                                              & 32010                        & 17456                        & $>10^6$ & 2249                         & 2926                         & 11108                        & 8328                         & 2778                     & $>10^{6}$ & $>10^{6}$                                                            & \textbf{2541}                                                                           & 10175                                                                                      & 2038                                                                                   \\
Seq-Elim.                                        & 10824                        & 17514                        & 5899                               & 4440                         & 16590                        & 6881                         & 17937                        & 12851                    & 48068                               & 38554                                                                                          & 41037                                                                                   & $>10^{6}$                                                        & 9046                                                                                   \\
PL                                               & 7011                         & 18513                        & 4774                               & 4618                         & 7859                         & 17049                        & 15215                        & 8037                     & 13156                               & \textbf{5682}                                                                                  & 60031                                                                                   & $>10^{6}$                                                        & 3871                                                                                   \\
Knockout                                         & 3415                         & 7889                         & 4723                               & 3444                         & 5104                         & 5809                         & 5956                         & 3134                     & 3777                                & 8055                                                                                           & 7708                                                                                    & 17418                                                                                      & 4953                                                                                   \\
Sing. Elim.                                     & 4830                         & 6000                         & 5885                               & 5340                         & 6953                         & 6465                         & 6453                         & 6000                     & 9000                                & 12940                                                                                          & 15000                                                                                   & 55900                                                                                      & 9045                                                                                   \\ \hline
RUCB                                             & 3125                         & 5697                         & 3329                               & 1636                         & \textbf{1655}                & 4536                         & 6222                         & 2732                     & 5617                                & 19024                                                                                          & 10924                                                                                   & 41149                                                                                      & 1647                                                                                   \\
RCS                                              & 2442                         & \textbf{3924}                         & 3370                               & 1537                         & 2662                         & 3867                         & 5296                         & \textbf{1816}                     & \textbf{4606}                       & 12678                                                                                          & 7263                                                                                    & 34709                                                                                      & 1903                                                                                   \\
RMED                                            & \textbf{2028}                & {5113}                & \textbf{1612}                      & \textbf{864}                 & 1707                         & \textbf{1929}                & \textbf{4047}                & {2093}            & 5647                                & 9364                                                                                           & 3753                                                                                    & \textbf{24132}                                                                             & \textbf{1162}       \\  \hline 
SAVAGE                                           & 10289                        & 18016                        & 6639                               & 2393                         & 2675                         & 12806                        & 12115                        & 5767                     & 22959                               & 39208                                                                                          & 41493                                                                                   & 255208                                                                                     & 4733                                                                                   \\
CCB                                              & 7017                         & 11267                        & 5389                               & 2884                         & 4092                         & 11548                        & 10905                        & 4386                     & 10020                               & 21392                                                                                          & 16960                                                                                   & 87138                                                                                      & 2518                                                                                   \\
DTS                                              & 10089                        & 9214                         & 8618                               & 4654                         & 4850                         & 13317                        & 16473                        & 4355                     & 11530                               & 18199                                                                                          & 19940                                                                                   & 170467                                                                                     & 1354                                                                                   \\
DTS++                                            & 7626                         & 9483                         & 5532                               & 2729                         & 6465                         & 9394                         & 14926                        & 9284                     & 17774                               & 31562                                                                                          & 15065                                                                                   & 52606                                                                                      & 6284                                                                                   \\ \Xhline{3\arrayrulewidth}
                                                                  
\end{tabular}}
\caption{Annotation complexity of 13 dueling bandit algorithms along with the uniform exploration algorithm on 13  datasets spanning 5 NLG tasks}
\label{tab:results_dueling_bandit_algorithms_appendix}
\end{table*}
\begin{table*}[]
\centering
\resizebox{0.999\textwidth}{!}{
\begin{tabular}{l|ccc|ccc|ccc|ccc|ccc|ccc}
\Xhline{3\arrayrulewidth}
\multicolumn{1}{l|}{\multirow{2}{*}{Metrics}} & \multicolumn{3}{c|}{\begin{tabular}[c]{@{}c@{}}WMT\\ (Micro Average)\end{tabular}}                                               & \multicolumn{3}{c|}{\begin{tabular}[c]{@{}c@{}}Grammarly\\ (Micro Average)\end{tabular}}                                         & \multicolumn{3}{c|}{\begin{tabular}[c]{@{}c@{}}CoNLL-2014\\  Shared Task\end{tabular}}                                           & \multicolumn{3}{c|}{\begin{tabular}[c]{@{}c@{}}E2E NLG\\  Challenge\end{tabular}}                                                & \multicolumn{3}{c|}{ParaBank}                                                                                                    & \multicolumn{3}{c}{TLDR OpenAI}                                                                                                 \\ \cline{2-19} 
\multicolumn{1}{l|}{}                         & \multicolumn{1}{c|}{Linear} & \multicolumn{1}{c|}{BTL} & \multicolumn{1}{c|}{\begin{tabular}[c]{@{}c@{}}BTL\\ Log.\end{tabular}} & \multicolumn{1}{c|}{Linear} & \multicolumn{1}{c|}{BTL} & \multicolumn{1}{c|}{\begin{tabular}[c]{@{}c@{}}BTL\\ Log.\end{tabular}} & \multicolumn{1}{c|}{Linear} & \multicolumn{1}{c|}{BTL} & \multicolumn{1}{c|}{\begin{tabular}[c]{@{}c@{}}BTL\\ Log.\end{tabular}} & \multicolumn{1}{c|}{Linear} & \multicolumn{1}{c|}{BTL} & \multicolumn{1}{c|}{\begin{tabular}[c]{@{}c@{}}BTL\\ Log.\end{tabular}} & \multicolumn{1}{c|}{Linear} & \multicolumn{1}{c|}{BTL} & \multicolumn{1}{c|}{\begin{tabular}[c]{@{}c@{}}BTL\\ Log.\end{tabular}} & \multicolumn{1}{c|}{Linear} & \multicolumn{1}{c|}{BTL} & \multicolumn{1}{c}{\begin{tabular}[c]{@{}c@{}}BTL\\ Log.\end{tabular}} \\ \hline
Chrf                     & 62.6         & 62.0      & 62.6                                                    & 75.7           & 75.3        & 75.9                                                      & 78.4          & 78.3        & 78.4                                                     & 47.4        & 48.8      & 48.3                                                    & 66.1   & 66.1 & 66.1                                               & 34.2   & 35.4 & 35.4                                               \\ 
Bleu-4                   & 41.5         & 53.4      & 41.5                                                    & 73.2           & 73.0        & 73.2                                                      & 78.9          & 78.7        & 78.9                                                     & 45.0        & 39.0      & 50.1                                                    & 63.8   & 63.2 & 63.8                                               & 42.8   & 44.0 & 42.8                                               \\ 
Rouge-L                  & 60.7         & 60.0      & 60.7                                                    & 73.5           & 73.6        & 73.6                                                      & 78.0          & 78.0        & 78.0                                                     & 44.6        & 43.8      & 50.2                                                    & 64.3   & 64.3 & 64.3                                               & 43.3   & 43.3 & 43.3                                               \\ \hline
Emb. Avg.                & 56.5         & 59.1      & 57.5                                                    & 70.1           & 70.3        & 71.5                                                      & 76.0          & 76.7        & 77.0                                                     & 49.8        & 51.6      & 51.8                                                    & 64.9   & 64.9 & 64.9                                               & 38.2   & 38.2 & 38.2                                               \\ 
Greedy Match             & 59.5         & 59.8      & 59.9                                                    & 68.1           & 68.4        & 68.2                                                      & 77.7          & 77.4        & 77.7                                                     & 46.5        & 48.8      & 48.9                                                    & 64.7   & 64.7 & 64.5                                               & 43.1   & 43.1 & 43.1                                               \\ 
Vector Extr              & 59.4         & 59.5      & 59.3                                                    & 66.0           & 66.9        & 66.5                                                      & 76.3          & 76.7        & 76.7                                                     & 44.9        & 46.2      & 49.1                                                    & 63.7   & 63.7 & 63.7                                               & 47.4   & 47.1 & 48.1                                               \\ \hline
Bertscore                & 65.9         & 66.2      & 65.9                                                    & 77.4           & 77.2        & 77.4                                                      & 82.0          & 81.5        & 82.0                                                     & 45.9        & 49.3      & 50.1                                                    & 68.1   & 68.1 & 68.1                                               & 44.5   & 44.4 & 44.5                                               \\ 
Laser                    & 65.3         & 65.1      & 65.3                                                    & 75.1           & 73.0        & 75.1                                                      & 78.0          & 76.4        & 78.0                                                     & 47.2        & 49.9      & 50.5                                                    & 67.0   & 67.0 & 67.0                                               & 35.4   & 35.4 & 35.4                                               \\ 
MoverScore               & 66.1         & 66.5      & 66.1                                                    & 74.7           & 70.9        & 73.0                                                      & 80.6          & 79.6        & 80.3                                                     & 50.1        & 49.3      & 50.4                                                    & 68.0   & 68.0 & 67.8                                               & 40.7   & 40.7 & 40.7                                               \\ \hline
Bleurt                   & 68.2         & 67.5      & 68.2                                                    & 77.1           & 76.6        & 76.0                                                      & 81.5          & 81.5        & 80.8                                                     & 48.1        & 50.4      & 50.4                                                    & 67.7   & 67.7 & 67.7                                               & 42.5   & 42.5 & 42.3                                               \\ \hline
Electra                  & \multicolumn{3}{c|}{65.7}                                                          & \multicolumn{3}{c|}{74.0}                                                                & \multicolumn{3}{c|}{81.6}                                                              & \multicolumn{3}{c|}{54.3}                                                         & \multicolumn{3}{c|}{81.7}                                          & \multicolumn{3}{c}{-}                                             \\ \Xhline{3\arrayrulewidth}
\end{tabular}}
\caption{Sentence-level accuracy of direct assessment metrics with linear, BTL, and BTL-logistic probability models and our trained Electra metric in predicting the comparison outcome}
\label{tab:results_metrics_appendix}
\end{table*}
\begin{figure*}
    \begin{center}
        \scalebox{.45}{\input{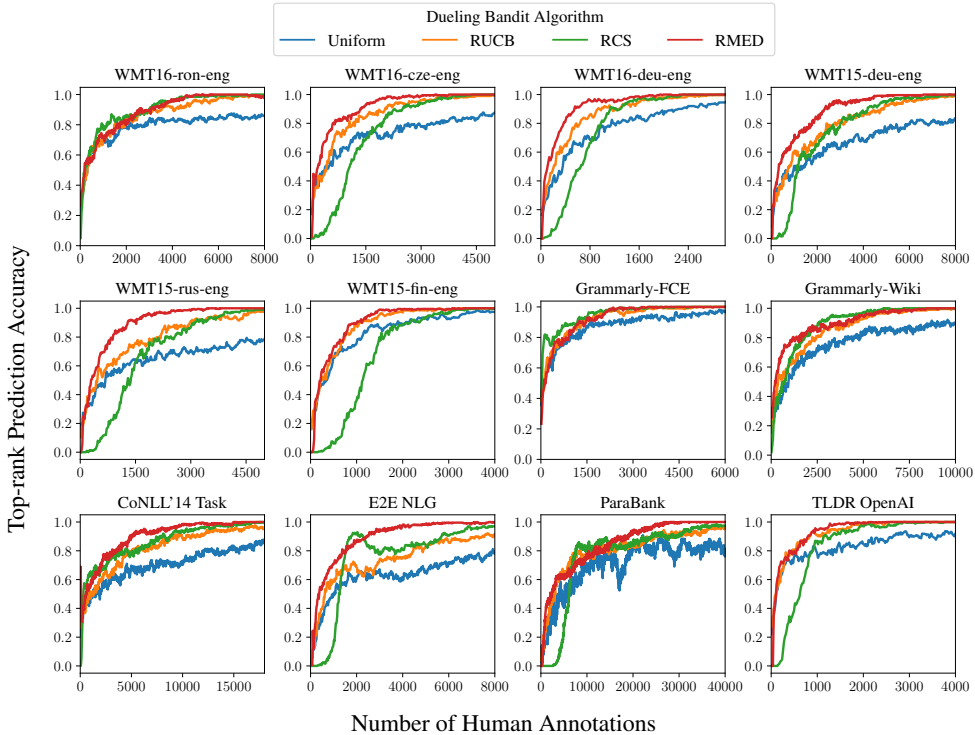}}
    \end{center}
    \caption{Top-rank prediction accuracy as a function of the number of human annotations for (model-free) Uniform exploration and RUCB, RCS, and RMED dueling bandit algorithms on 12 NLG datasets}
    \label{fig:top_ranked_bandits_appendix}
\end{figure*}
\begin{figure*}
    \begin{center}
        \scalebox{.45}{\input{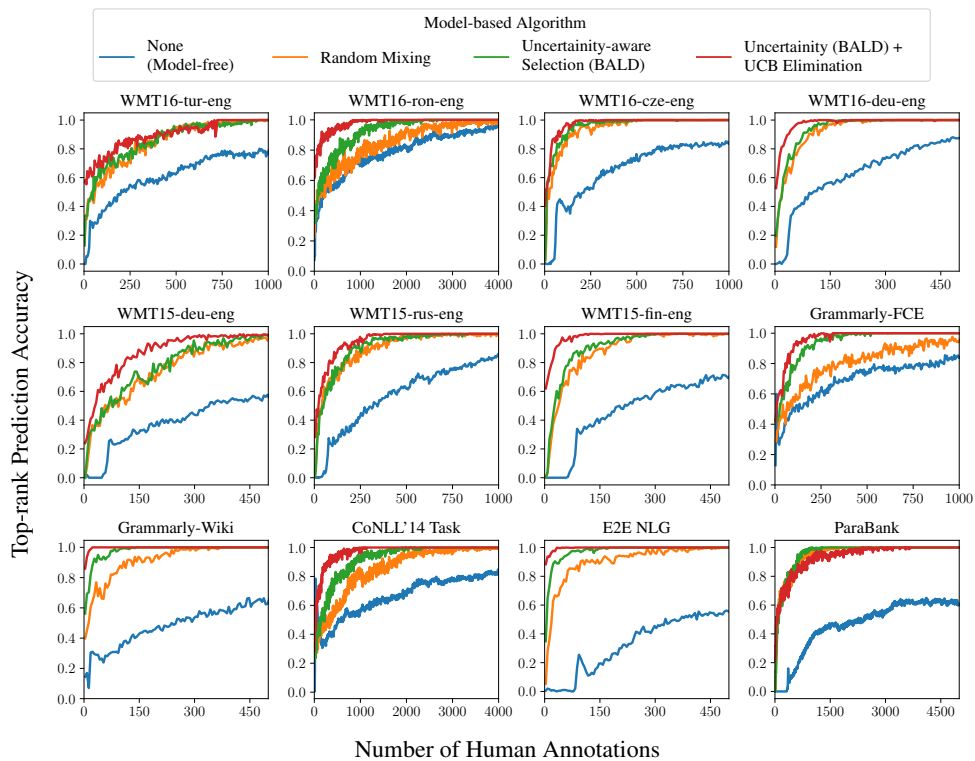}}
    \end{center}
    \caption{Top-rank prediction accuracy as a function of the number of human annotations for various model-based dueling bandit algorithms with RMED and Electra metric on 12 NLG datasets}
    \label{fig:top_ranked_model_based_appendix}
\end{figure*}
\section{Additional Results}
\subsection{Effect of number of NLG systems}
\label{appendix:num_of_nlg_systems}
In figure \ref{fig:number_of_nlg_systems_rand_mix}, we compare the variations in annotation complexity of Random Mixing (with Electra metric) using uniform exploration and dueling bandit algorithms. Similar to the model-free case discussed in section \ref{subsec:num_nlg_systems}, the annotation complexity of uniform exploration grows as $O(k^2)$ but the annotation complexity only varies as $O(k)$ for RMED, RCS, and RUCB dueling bandit algorithms
\subsection{Results of Dueling Bandit Algorithms}
\label{appendix:results_dueling_bandit_algorithms}
We report the annotation complexity of all 13 dueling bandit algorithms on 13 evaluation datasets in table \ref{tab:results_dueling_bandit_algorithms_appendix}. In figure \ref{fig:top_ranked_bandits_appendix}, we show the top-rank prediction accuracy as a function of the number of human annotations for various dueling bandit algorithms on all the datasets, other than WMT 16 tur-eng, which is separately depicted in figure \ref{fig:toprank_preidction_accuracy}. 
\subsection{Performance of Evaluation Metrics}
\label{appendix:results_automatic_metrics}
In table \ref{tab:results_metrics_appendix}, we report the sentence-level accuracy in predicting the comparison outcome for 10 direct assessment metrics using three probability models along with the trained pairwise metric (Electra). We observe that there is little variation in performance across the three probability models. To further illustrate this, we plot the accuracy on the WMT datasets in figure \ref{fig:probability_models} and observe that the performance is largely similar across Linear, BTL, and BTL-logistic models. 

\subsection{Model-based Algorithms}
\label{appendix:model_based_algo}
In figure \ref{fig:top_ranked_model_based_appendix}, we show the top-rank prediction accuracy as a function of the number of human annotations for various model-based algorithms using the Electra metric with RMED. We observe that Random Mixing and Uncertainty-aware Selection (BALD) algorithms have significantly higher prediction accuracy than model-free RMED for any given number of human annotations. Further, when we use UCB Elimination with Uncertainty-aware Selection, we observe the highest top-rank prediction accuracy for any given number of annotations. 
. 
\end{document}